\documentclass[sigconf]{acmart}

\AtBeginDocument{%
  \providecommand\BibTeX{{%
    \normalfont B\kern-0.5em{\scshape i\kern-0.25em b}\kern-0.8em\TeX}}}

\usepackage{graphicx}
\usepackage{amsmath}
\usepackage{amsthm}
\usepackage{multirow}
\usepackage{algorithm}
\usepackage{subfig}
\usepackage{algorithmic}
\usepackage{bm}
\usepackage{amssymb}%
\usepackage{bm}
\newcommand{\tabincell}[2]{\begin{tabular}{@{}#1@{}}#2\end{tabular}}
\copyrightyear{2020} 
\acmYear{2020} 
\setcopyright{acmcopyright}
\acmConference[MM '20]{Proceedings of the 28th ACM International Conference on Multimedia}{October 12--16, 2020}{Seattle, WA, USA}
\acmBooktitle{Proceedings of the 28th ACM International Conference on Multimedia (MM '20), October 12--16, 2020, Seattle, WA, USA}
\acmPrice{15.00}
\acmDOI{10.1145/3394171.3414026}
\acmISBN{978-1-4503-7988-5/20/10} 
\newcommand\blfootnote[1]{% 
\begingroup 
\renewcommand\thefootnote{}\footnote{#1}% 
\addtocounter{footnote}{-1}% 
\endgroup 
}

\acmSubmissionID{mmfp0241}

\begin{document}
\fancyhead{}
\title{Jointly Cross- and Self-Modal Graph Attention Network for Query-Based Moment Localization}
% Graph based Cross- and Self-modality Attentive Networks for Moment Retrieval with Enhanced Sentence Understanding

\author{Daizong Liu$^\dagger$}
\affiliation{%
  \institution{School of Electronic Information and Communication, Huazhong University of Science and Technology}
%   \city{Wuhan}
%   \country{China}
  }
\email{dzliu@hust.edu.cn}

\author{Xiaoye Qu$^\dagger$}
\affiliation{%
  \institution{Huawei Cloud, \\Hangzhou, China}
%   \city{Wuhan}
%   \country{China}
  }
\email{quxiaoye@huawei.com}

\author{Xiao-Yang Liu}
\affiliation{%
  \institution{Department of Electrical Engineering, Columbia University}
%   \country{United States}
  }
\email{xl2427@columbia.edu}

\author{Jianfeng Dong}
\affiliation{%
  \institution{School of Computer and Information Engineering, Zhejiang Gongshang University}
%   \city{Hangzhou}
%   \country{China}
  }
\email{dongjf24@gmail.com}

\author{Pan Zhou$^*$}
\affiliation{%
  \institution{The Hubei Engineering Research Center on Big Data Security, School of Cyber Science and Engineering, Huazhong University of Science and Technology}
%   \city{Wuhan}
%   \country{China}
  }
\email{panzhou@hust.edu.cn}

\author{Zichuan Xu}
\affiliation{%
  \institution{School of Software, Dalian University of Technology}
%   \city{Dalian}
%   \country{China}
  }
\email{z.xu@dlut.edu.cn}

% \author{
%     Anonymous submission
% }
% \affiliation{
%     Paper ID 241
% }


\begin{abstract}
Query-based moment localization is a new task that localizes the best matched segment in an untrimmed video according to a given sentence query.
% \yanglet{"important, difficult, challenging?"}
In this localization task, one should pay more attention to thoroughly mine visual and linguistic information. 
% while previous works fail to develop a comprehensively framework for this challenging task.
% Previous works generally focus on integrating multi-modal features by cross-modal attention mechanisms, while fail to additionally consider the self-modal relation that contributes to distinguishing and correlating the sequential activities. 
% However, learning to effective capture both cross- and self-modal relations is at the heart of this challenging task. 
To this end, we propose a novel \textbf{C}ross- and \textbf{S}elf-\textbf{M}odal \textbf{G}raph \textbf{A}ttention \textbf{N}etwork (CSMGAN) that recasts this task as a process of iterative messages passing over a joint graph.
Specifically, the joint graph consists of \textbf{C}ross-\textbf{M}odal relation \textbf{G}raph (CMG) and \textbf{S}elf-\textbf{M}odal relation \textbf{G}raph (SMG), where frames and words are represented as nodes, and the relations between cross- and self-modal node pairs are described by an attention mechanism. Through parametric message passing,
% Specifically, CSMGAN builds successive \textbf{C}ross-\textbf{M}odal interaction \textbf{G}raph (CMG) and \textbf{S}elf-\textbf{M}odal relation \textbf{G}raph (SMG) for both cross- and self-modal information capturing. 
% In each joint graph layer, frames and words are represented as nodes, and the relations between cross- and self-modal node pairs denoted as edges are described by a differentiable attention mechanism. Through parametric message passing,
CMG highlights relevant instances across video and sentence, and then SMG models the pairwise relation inside each modality for frame (word) correlating. With multiple layers of such a joint graph, our CSMGAN is able to effectively capture high-order interactions between two modalities, thus enabling a further precise localization. 
Besides, to better comprehend the contextual details in the query, we develop a hierarchical sentence encoder to enhance the query understanding. Extensive experiments on two public datasets 
% (Activity Caption, TACoS, Charades-STA and DiDeMo) 
demonstrate the effectiveness of our proposed model, and GCSMAN significantly outperforms the state-of-the-arts.
The code is available at \href{https://github.com/liudaizong/CSMGAN}{https://github.com/liudaizong/CSMGAN}.
% Extensive experiments on public datasets demonstrate the effectiveness of our proposed model.
% and our GCSMAN outperforms the state-of-the-arts with clear margins.

% Query-based moment retrieval is a new task to localize the most relevant segment in an untrimmed video according to a given sentence query. Previous works generally focus on designing effective cross-modal attention mechanisms to explore the potential interaction between video and query. However, they fail to exploit the self-modal relation contributing to distinguishing and correlating sequential activities for temporal localization.
% In this paper, we propose a novel cross- and self-modal graph attention network (CSMGAN) for this challenging task, which consists of multi-layer jointly \textbf{c}ross-\textbf{m}odal interaction \textbf{g}raph (CMG) and \textbf{s}elf-\textbf{m}odal relation \textbf{g}raph (SMG). Specifically, in each layer, CMG highlights relevant components across the video and sentence. Subsequently, SMG models the pairwise relation inside each modality for frame (word) association. The stacking of multi-layer such jointly cross- and self-modal graph can capture high-order interactions between two modalities, thus enabling a further precise localization. Besides, to better mine the guiding clues within the query, we develop a hierarchical structure to enhance the sentence understanding. Extensive experiments on two public datasets (Activity Caption and TACoS) demonstrate the effectiveness of our proposed model, and our GCSMAN outperforms the state-of-the-arts with clear margins.
\end{abstract}



\maketitle

\blfootnote{$^\dagger$Equal Contribution.}
\blfootnote{$^*$This work was supported in part by the National Natural Science Foundation of China (No. 61972448, No. 61902347), and Zhejiang Provincial Natural Science Foundation (No. LQ19F020002). Corresponding author: Pan Zhou.}

\vspace{-20pt}
\section{Introduction}
% Video is an increasingly popular topic in multimedia information understanding \cite{regneri2013grounding,yuan2016temporal,gavrilyuk2018actor,feng2018video,feng2019spatio}. As most videos contain activities of interest with complicated background content, locating the specific activity segment is the key to better analyse the video. Recently, the task of query-based moment retrieval in video has gained great interests from the computer vision \cite{gao2017tall,anne2017localizing}.
Localizing activities in videos  \cite{regneri2013grounding,yuan2016temporal,gavrilyuk2018actor,feng2018video,feng2019spatio} is an important topic in multimedia information retrieval.
However, in realistic scenario, YouTube videos normally contain complicated background contents, and cannot be directly indicated by a pre-defined list of action classes.
% leading the precisely localization of interested activities much harder.
% One of the practical scenario is that people could search for the segment they want to watch with keywords among YouTube videos. However, most videos contain activities of interest with complicated background contents, and cannot be directly indicated by a pre-defined list of action classes. 
To address this problem, query-based moment localization is proposed recently \cite{gao2017tall,anne2017localizing} and attracts increasing interests from the multimedia community \cite{liu2018cross,zhang2019exploiting}.
It aims to ground the most relevant video segment according to a given sentence query. This task is challenging because most part of video contents are irrelevant to the query while only a short segment matches the sentence. Therefore,
video and sentence information need to be deeply incorporated to distinguish the fine-grained details of different video segments and perform accurate segment localization. 

% \begin{figure}[t]
%     \centering
%     \includegraphics[width=0.48\textwidth]{figure/introduction6.pdf}
%     \caption{CSMGAN represents words and frames (dark yellow/green) as nodes, and the relations between node pairs as edges (all lines) captured by attention mechanism. It builds successive cross-modal graph for relevant components highlighting (bright yellow/green) and self-modal graph for sequential elements correlating (red line).}
%     \label{fig:introduction}
%     \vspace{-8pt}
% \end{figure}

\begin{figure}[t]
    \centering
    \includegraphics[width=0.48\textwidth]{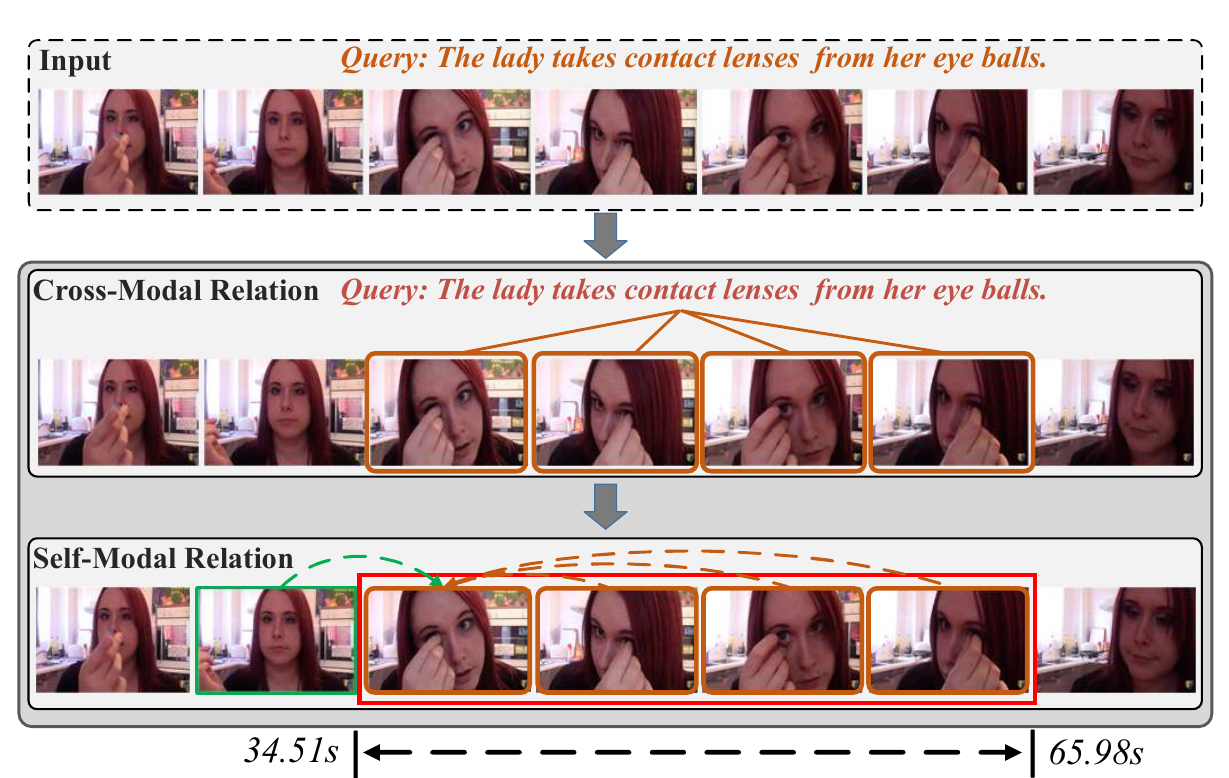}
    \caption{Given a query and an untrimmed video, CSMGAN considers cross-modal relation for highlighting relevant instances (brown rectangles), and self-modal relation for correlating sequential elements (red rectangle) and distinguishing components near the boundary (green rectangle).}
    \label{fig:introduction}
    \vspace{-15pt}
\end{figure}

Most existing methods \cite{liu2018attentive,chen2018temporally,zhang2019cross,yuan2019find,yuan2019semantic} for this task focus on learning the cross-modal relations between video and sentence. Specifically, they develop attention based interaction mechanisms to enhance the video representation with sentence information. 
% Liu \textit{et al.} \cite{liu2018attentive} design a memory network on query representation to emphasize the visual features mentioned in the sentence.
% Yuan \textit{et al.} \cite{yuan2019semantic} introduce the sentence information as a critical prior to compose and correlate the video contexts.
% Yuan \textit{et al.} \cite{yuan2019find} utilizes a co-attention mechanism to generate query-guided video representation.
% Some works \cite{chen2018temporally,zhang2019cross,yuan2019find} also develop attention based interaction mechanisms to enhance the video representation with sentence information. 
Meanwhile, few algorithms \cite{chen2017msrc,zhang2019cross} attempt to learn the self-modal relations. For example, 
% Chen \textit{et al.} \cite{chen2017msrc} devise a self-interactor to perform cross-frame matching. 
% Wang \textit{et al.} \cite{wang2019temporally} leverage attention mechanism to explore contextual integration, and 
Zhang \textit{et al.} \cite{zhang2019cross} leverage self-attention to capture long-range semantic dependencies just in video encoding. 
% However, the effective interaction of multi-modal representation depending on successively capturing both cross- and self-modal relations are never jointly investigated in a joint framework for this task. 
% However, for multi-modal localization problem, the self-modal relation within each modality is complementary to the cross-modal relation, 
However, the cross- and self-modal relations are never jointly investigated in a joint framework for this task. 
As shown in Figure \ref{fig:introduction}, for the video modality, each frame should not only obtain information from its associated words in the query to highlight relevant frames (brown rectangles), but also need to correlate these highlighted frames to infer the sequential activity (red rectangle). At the same time, as the adjacent frame (green rectangle) near the boundary shows different visual appearance, such self-modal relation also contributes to distinguishing the segment boundaries for more precise localization.    
% such self-modal relation also contributes to distinguishing visually similar frames near the segment boundaries for more precise localization. 
Similarly, for the query modality, a better understanding of sentence can be acquired in conjunction with both frames and other words. Such cases motivate us to propose a joint framework for modelling both cross- and self-modal relations.
% However, the cross-modal relation is as important as the self-modal relation, and they are never jointly investigated in a joint framework for this task. We argue that, for multi-modal localization problem, the self-modal relations within each modality is complementary to the cross-modal relations, which are mostly ignored by existing methods. For instance, for the video modality, each frame should obtain information not only from its associated words in the query but also from related frames to infer the sequential activity mentioned by the query. The self-modal relation also contributes to distinguishing and correlating visually similar frames near the segment boundaries for more precise localization. For the query modality, better understanding of sentence can be acquired in conjunction with other words. Such cases motivate us to propose a joint framework for modelling both cross- and self-modal relations.

% as graph neural network (GNN) takes advantage of aggregating the information with node-wise correlation establishment, 
In this paper, we develop a novel cross- and self-modal graph attention network (CSMGAN) for query-based moment localization, which recasts this task as an end-to-end, message passing based joint graph information fusion procedure. The joint graph consists of a cross-modal relation graph (CMG) and a self-modal relation graph (SMG), and represents both video frames and sentence words as nodes. Specially, in each joint graph layer, CMG first establishes the edges between each word-frame pair for cross-modal information passing, where the directed pair-wise relations are efficiently captured by a heterogeneous attention mechanism. Subsequently, SMG is designed to capture the complex self-modal relations by establishing the edges within each modality. 
The combination of CMG and SMG makes it possible to obtain more contextual representations by correlating highlighted cross-modal instances with sequential elements.
Moreover, by stacking multiple layers to recursive propagate messages over the joint graph, our CSMGAN can capture higher-level relationships among multi-modal representations, and comprehensively integrates the localization information for precise moment retrieval.

Besides, traditional methods \cite{zhang2019man,yuan2019semantic,wang2019temporally,mithun2019weakly,yuan2019find,chen2019semantic} adopt RNN for sentence query embedding. 
% We also argue that there requires a more detailed structure for sentence understanding. To better comprehend the sentence,
However, they fail to explicitly consider the multi-granular textual information, such as specific phrases which are crucial to understanding the sentence. To capture the fine-grained query representations, we build a hierarchical structure to understand the query at three levels: word-, phrase- and sentence-level. These hierarchical representations are then merged to stand for a more informative understanding of the sentence query. 
% At word-level, we embed each word with word2vec \cite{pennington2014glove} encoder. At the phrase-level, 1D convolution layers are used to capture the information contained in unigrams, bigrams and trigrams. Specifically, we convolve word-level representations with temporal filters of varying support, and then combine the various n-gram responses by pooling them into a single phrase-level representation. At sentence-level, we apply a RNN to encode the phrase-level representation. At last, we fuse these three-level representations to generate final contextual sentence embedding.

In summary, the main contributions of our work are:
\vspace{-5pt}
\begin{itemize}
    \item We present a cross- and self-modal graph attention network (CSMGAN), which is made up of cross- and self-modal graph for localizing desired moments. To the best of our knowledge, it is the first time that a joint framework is proposed to consider both cross- and self-modal relations for query-based moment localization. 
    % \item We design a jointly cross- and self-modal graph attention network (CSMGAN) for multi-modal integration by interleaving cross- and self-modal feature association. It is the first time that a joint framework is proposed to integrate both cross- and self-modal information for query-based moment retrieval task.
    \item We design a hierarchical structure to capture the fine-grained sentence representation at three different levels: word-level, phrase-level and sentence-level. 
    % These three-level features are then combined to generate the final sentence representation.
    \item We conduct experiments on Activity Caption and TACoS datasets and our CSMGAN outperforms the state-of-the-arts with clear margins.
    % our CSMGAN achieves the state-of-the-art performances. 
    % Ablation studies and qualitative analysis demonstrate the effectiveness of 
    % \yanglet{our proposed framework, should verify the combination of cross- and self-modal base scheme, right?}.
\end{itemize}

% The rest of the paper is organized as follows. Section 2 reviews the related work. Section 3 details our proposed CSMGAN model. We present the experimental results in Section 4, followed by the conclusion in Section 5.

%%%%%%%%%%%%%%%%%%%%%%%%%%%%%%%%%%%%%%%%%%%
\vspace{-10pt}
\section{Related works}
% In this section, we briefly review some related works on image and video localization tasks and graph neural networks.

\noindent \textbf{Query-based localization in images.} Early works of localization task mainly focus on localizing the image region corresponding to a language query. They first generate candidate image regions using image proposal method \cite{ren2015faster}, and then find the matched one with respect to the given query. Some works \cite{mao2016generation,hu2016natural,rohrbach2016grounding} try to extract target image regions based on description reconstruction error or probabilities. There are also several studies \cite{yu2016modeling,chen2017query,chen2017msrc,zhang2018grounding} considering incorporating contextual information of region-phrase relationship into the localization model. \cite{wang2016structured} further models region-region and phrase-phrase structures. Some other methods exploit attention modeling in queries, images, or object proposals \cite{endo2017attention,deng2018visual,yu2018mattnet}.

\begin{figure*}[t]
    \centering
    \includegraphics[width=1.0\textwidth]{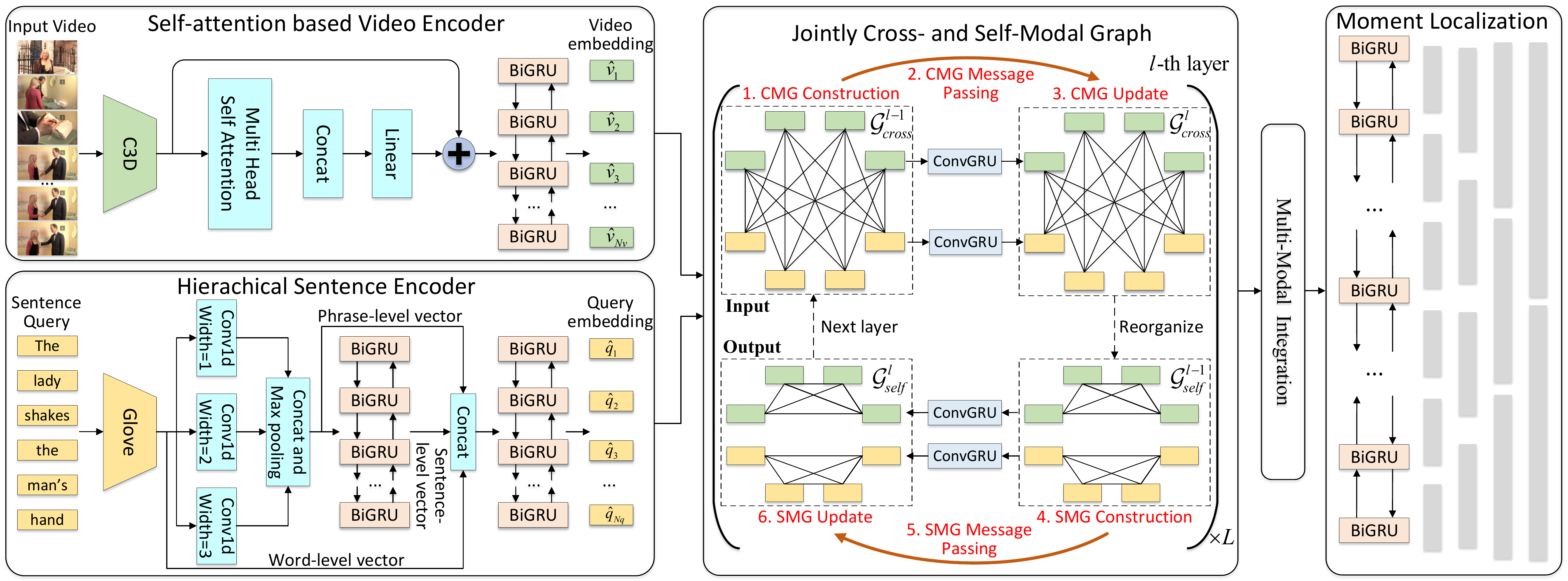}
    \caption{Illustration of our proposed CSMGAN. We first utilize a self-attention based video encoder and a hierarchical sentence encoder to extract corresponding features. Then, a jointly cross- and self-modal graph is devised for multi-modal interaction. In the joint graph, words and frames both represented as nodes first construct CMG to mine cross-modal relations and update their states through ConvGRU. Then the nodes are reorganized as SMG to model the self-modal relationships and updated for the next layer graph input. At last, we conduct multi-modal integration and perform moment localization.}
    \label{fig:pipeline}
    \vspace{-8pt}
\end{figure*}

\noindent\textbf{Query-based moment localization in videos.}
Different from traditional video retrieval \cite{dong2018predicting, dong2019dual}, it is a new task introduced recently \cite{gao2017tall,anne2017localizing}, which aims to localize the most relevant video segment from a video with text descriptions. Traditional methods \cite{liu2018attentive,gao2017tall} sample candidate segments from a video first, and subsequently integrate query with segment representations via a matrix operation. 
% These methods lack a comprehensively structure for effective multi-modal features interaction. 
To further mine the cross-modal interaction more effectively, some works \cite{xu2019multilevel,chen2019semantic,ge2019mac,zhang2019learninga} integrate the sentence representation with those video segments individually, and then evaluated their matching relationships.
% Liu \textit{et al.} \cite{liu2018attentive} design a memory attention mechanism on query sentence to emphasize the visual features mentioned in the sentence.
For instance, Xu \textit{et al.} \cite{xu2019multilevel} introduce a multi-level model to integrate visual and textual features earlier and further re-generate queries as an auxiliary task.
% Chen \textit{et al.} \cite{chen2019semantic} propose to generate query-specific proposals as candidate segments. 
Chen \textit{et al.} \cite{chen2018temporally} capture the evolving fine-grained frame-by-word interactions between video and query to enhance the video representation understanding.
Recently, other works \cite{chen2018temporally,wang2019temporally,zhang2019cross,zhang2019man,yuan2019semantic,mithun2019weakly} propose to directly integrate sentence information with fine-grained video clip, and predict the temporal boundary of the target segment by gradually merging the fusion feature sequence over time.
% Wang \textit{et al.} \cite{wang2019temporally} aggregate contextual information by explicitly modeling the relationship between the current element and its neighbors.
Zhang \textit{et al.} \cite{zhang2019man} model relations among candidate segments produced from a convolutional neural network with the guidance of the query information.
To modulate temporal convolution operations, Yuan \textit{et al.}\cite{yuan2019semantic} and Mithun \textit{et al.} \cite{mithun2019weakly} introduce the sentence information as a critical prior to compose and correlate video contents. 
Although these methods achieve relatively superior performances by capturing cross-modal information, they ignore to utilize the self-modal relation which is complementary to the cross-modal relation.
% However, previous works have several limitation for better cross-modal interaction. First, these methods all adopt frame-by-word interactions to fuse the cross-modal information. That is, they only corporate the sentence meaning for better video contexts understanding, while ignoring to mine the deep correlation from video frame to sentence query. Semantic sentence information enhanced by video-guided composing can also help to enhance back for video understanding. Second, most of previous works generally capture the temporal information among frames or words by only using bi-directional RNN. It lacks full exploring their intra-dependencies. 
Different from them, we propose a cross- and self-modal graph attention network to jointly consider both cross- and self-modal relations. The successive cross- and self-modal graphs enable our model to capture much higher-level interactions.

\noindent \textbf{Graph neural networks.}
Graph neural network (GNN) \cite{scarselli2008graph} is an extension for recursive neural networks and random walk based models for graph structured data. As a follow-up work,  Gilme \textit{et al.} \cite{gilmer2017neural} further adapt GNN to sequential outputs with a learnable message passing module.
% Generally, for each node in a graph, the updating process includes two steps: node message aggregation and node state update. To be specific, each node first aggregates messages from its neighbors into a single message and then updates the node state itself with such message. 
As GNN is wildly used in sequential information processing, in this paper, we design a novel GNN module for cross- and self-modal relations mining. Different from original GNN, we represent edge weights by an attention mechanism and aggregate messages with a gate function. Moreover, we utilize a ConvGRU \cite{ballas2015delving} layer for node state updating.

\vspace{-7pt}
\section{The proposed CSMGAN framework}
% In this section, we present our method CSMGAN, as illustrated in Figure \ref{fig:pipeline}. Given a video-sentence pair, we first utilize a hierarchical sentence encoder to embed the sentence representation, and a self-attention based module to extract the video representation.
% Then, to better interact multi-modal features, we propose an unified cross- and self-modal graph attention network for jointly considering both cross- and self-modal relations. In details, this graph contains two subgraphs: a cross-modal interaction graph is utilized to mine the correlations between the word-frame pairs for aggregating the interdepended information; and 
% % Then, to mine the correlations between the word-frame pairs, we propose a novel cross-modal graph for aggregating the interdepended information. 
% % After getting the video-aware sentence representation and sentence-aware video representation, 
% subsequently, a self-modal relation graph is adopted to further collect contextual grounding clues.  
% % we further merge them with a fusion function. 
% At last, we utilize a multi-modal fusion module for multi-modal representation fusion, and a moment retrieval module to localize the boundaries of target moments.

\subsection{Overview}
Given an untrimmed video $\bm{V}$ and a sentence query $\bm{Q}$, the task aims to determine the start and end timestamps $(s, e)$ of specific video segment referring to the sentence query. Formally, we represent the video as $\bm{V}=\{\bm{v}_t\}_{t=1}^{N_v}$ frame-by-frame, where $\bm{v}_i$ is the $i$-th frame in the video and $N_v$ is the total frame number. We also denote the given sentence query as $\bm{Q}=\{\bm{q}_n\}_{n=1}^{N_q}$ word-by-word, where $\bm{q}_n$ is the $n$-th word. 
With the training set $\{V,Q,(s,e)\}$, we aim to learn to predict the most relevant video segment boundary $(\hat{s}, \hat{e})$ which conforms to the sentence query information.

We present our method CSMGAN in Figure \ref{fig:pipeline}. 
First of all, a self-attention based video encoder and a hierarchical sentence encoder are utilized to extract contextual sentence and video embeddings.
Then, in order to better interact multi-modal features, we capture both cross- and self-modal relations by developing a jointly cross- and self-modal graph network.
Specially, in the joint graph, a cross-modal relation graph (CMG) establishes weighted edges between frame-word pairs to pass the message flows across the modalities. Following it, a self-modal relation graph (SMG) reorganizes the previous nodes and edges to model the relationships within each modality. With the self-relation complemented to the cross-relation, the joint graph can perform richer interaction of multi-modalities. Moreover, with multiple layers of such joint graph, the CSMGAN can capture higher-order relationships.
At last, the enhanced two modal representations are integrated to score different candidate video segments by a moment localization module.
% In details, the jointly graph contains two subgraphs: a cross-modal interaction graph and a self-modal relation graph which propagate attentive flows among cross-modality and self-modality, respectively. 
% At last, we utilize an integration module for multi-modal representation fusion, and devise a moment retrieval module to localize the boundaries of target moments.
% To deal with this task, we present our method CSMGAN as illustrated in Figure \ref{fig:pipeline}. We first utilize a hierarchical sentence encoder to embed the sentence representation, and a self-attention based video encoder to extract the video representation for informative contexts capturing. Both two encoders are detailed described in Section 3.1. 
% Then, in order to better interact multi-modal representations, we propose a jointly cross- and self-modal attentive graph in Section 3.2 for joint considering both cross- and self-modal relations. 
% % In details, the jointly graph contains two subgraphs: a cross-modal interaction graph and a self-modal relation graph which propagate attentive flows among cross-modality and self-modality, respectively. 
% Subsequently, we utilize a multi-modal integration module for multi-modal representation fusion in Section 3.3. Finally, in Section 3.4, we devise a moment retrieval module to localize the boundaries of target moments.
\vspace{-8pt}

\begin{figure*}[t]
    \centering
    \includegraphics[width=1.0\textwidth]{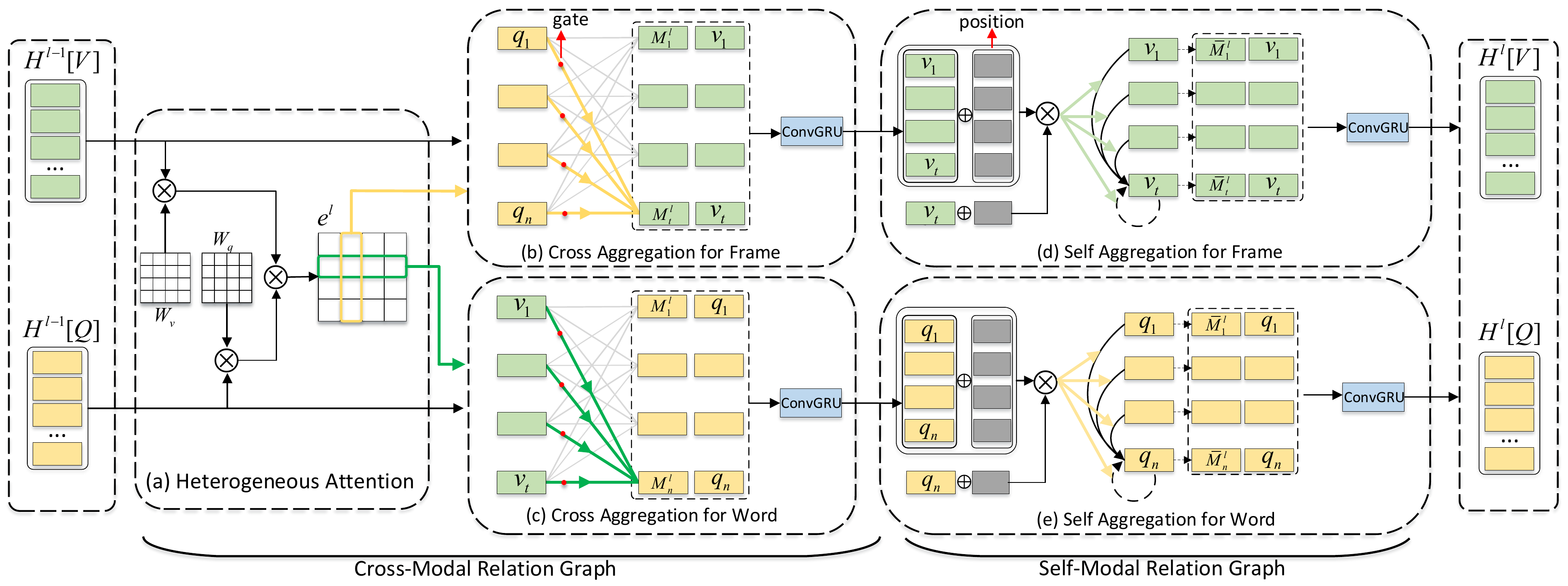}
    \caption{Illustration of our cross-modal relation graph and self-modal relation graph. Both two graphs compute attention matrices to stand for the attention weights on the corresponding edges. Each node aggregates the messages from its neighbor nodes in an edge-weighted manner and updates its state with both aggregated message and current state through ConvGRU. We apply a gate function in cross-modal graph and consider the temporal position for all nodes in self-modal graph.}
    \label{fig:edge}
    \vspace{-8pt}
\end{figure*}

\subsection{Video and Sentence Encoder}
\textbf{Video encoder.} 
Following \cite{zhang2019cross}, we first extract the frame-wise features by a pre-trained C3D network \cite{tran2015learning}, and then employ a self-attention \cite{vaswani2017attention} module to capture the long-range dependencies among video frames.
Considering the sequential characteristic in video, a bi-directional GRU \cite{chung2014empirical} is further utilized to capture the contextual information in time series. We denote the encoded representation as $\hat{\bm{V}}=\{\hat{\bm{v}}_t\}_{t=1}^{N_v} \in \mathbb{R}^{N_v \times d}$, where the $\hat{\bm{v}}_t \in \mathbb{R}^{1 \times d}$ is the feature of the $t$-th frame.\\
\textbf{Sentence encoder.} 
% Most previous works directly adopt recurrent neural networks to model natural language queries. 
% Although a syntactic relation \cite{zhang2019cross} is developed for better query semantic understanding, these works ignore a general way to capture the meaning of a sentence: firstly concatenate words to phrases, and then jointly reading the phrases to know the entire sentence. 
Most previous works generally adopt recurrent neural networks to model the contextual information for each word during the sentence encoding process. However, considering the query ``He continues playing the instrument", it is reasonable to focus on the phrase ``continues playing" instead of each single word to obtain more detailed temporal clues for precise localization. Therefore, to fully mine the guiding information, we develop a hierarchical structure with word-, phrase-, and sentence-level feature extracting for sentence query encoding.

We first generate the word-level features for the query by using the Glove word2vec embedding \cite{pennington2014glove}, and denote them as $\bm{Q}^w=\{\bm{q}_n^w\}_{n=1}^{N_q} \in \mathbb{R}^{N_q \times d_g}$, where $N_q$ is the number of words in the sentence and $d_g$ is the Glove embedding dimension. 
To discover the potential phrase-level features, we apply 1D convolutions on the word-level features with different window sizes. Specially, at each word location, we compute the inner product of the word feature vectors with convolution filters of three kinds of window sizes, which captures unigram, bigram and trigram features. 
% To capture the phrase-level features, we apply three 1D convolutions on the word-level features with different window sizes. Specially, at each word location, we compute the inner product of the word feature vectors with convolution filters of three kinds of window sizes, which denotes unigram, bigram and trigram features. 
% To further enlarge the convolution receptive field, we utilize the dilated convolution \cite{wang2018understanding} here. 
To maintain the sequence length after convolution process, we zero-pad the sequence vectors when convolution window size is larger than one. 
The output of the $n$-th word location with window size $k \in \{1,2,3\}$ is formulated as follows:
\begin{equation}
    \bm{q}_{n,k}^p = \text{tanh}(\text{Conv1d}(\bm{q}^w_{n:n+k-1}))  \in \mathbb{R}^{1 \times d_g},
\end{equation}
where $\text{Conv1d}(\cdot)$ operates on the windowed features with $d_g$ kernels. $\bm{q}_{n,k}^p$ is the phrase-level feature corresponding to $n$-th word location with window size $k$. To find the most contributed phrase at each word location, we then apply max-pooling to obtain the final phrase-level feature $\bm{Q}^p=\{\bm{q}_n^p\}_{n=1}^{N_q} \in \mathbb{R}^{N_q \times d_g}$ by:
\begin{equation}
    \bm{q}_n^p = \text{max}(\bm{q}_{n,1}^p, \bm{q}_{n,2}^p, \bm{q}_{n,3}^p), n \in \{1,2,...,N_q\}.
\end{equation}
After obtaining the phrase-level feature vector $\bm{Q}^p$, we encode them with a bi-directional GRU network to produce the sentence-level feature $\bm{Q}^s =\{\bm{q}_n^s\}_{n=1}^{N_q} \in \mathbb{R}^{N_q \times d_g}$. At last, we concat these three-level features and leverage another bi-directional GRU network to integrate them by:
% \begin{equation}
%     \bm{Q}' =  \in \mathbb{R}^{N_q \times 3d_g}
% \end{equation}
\begin{equation}
    \hat{\bm{Q}} = \text{Bi-GRU}(\text{Concat}[\bm{Q}^w,\bm{Q}^p,\bm{Q}^s]) \in \mathbb{R}^{N_q \times d}.
\end{equation}
Here the contextual query representation $\hat{\bm{Q}}=\{\hat{\bm{q}}_n\}_{n=1}^{N_q}$ is projected from the length $3d_g$ to $d$ to keep same dimension as video representation. 
% This hierarchical embedding structure performs like a human reading way where a sentence is comprehended by a word-phrase-sentence process. 
% And it can produce fine-grained features for better understanding the given sequence.
After the hierarchical embedding structure, the given query can obtain comprehensive understanding and provide robust representation for later localization. 

% After getting the video representation $\hat{\bm{V}}=\{\hat{\bm{v}}_t\}_{t=1}^{N_v}$ and query representation $\hat{\bm{Q}}=\{\hat{\bm{q}}_n\}_{n=1}^{N_q}$, 
% we first develop a cross-modality graph where each word (frame) is taken as a node and the connected edges only exist between the word-frame pairs. This graph is a directed graph, we denote it as $\mathcal{G}_{inter}=(\mathcal{V}_{inter}, \mathcal{E}_{inter})$, where $\mathcal{V}$ contains all nodes of both words and frames, edge $e_{(t,n)}=(\bm{v}_t, \bm{q}_n) \in \mathcal{E}_{inter}$ indicates the relation from video $\bm{v}_t$ to word $\bm{q}_n$ and vice versa. 
% Following by this graph, we then develop intra-modality graphs between word-word and frame-frame pairs. And we denote it as $\mathcal{G}_{intra}=(\mathcal{V}_{intra}, \mathcal{E}_{intra})$.
% The core idea of this module is to perform $L$ message propagation iterations over inter-intra graph $(\mathcal{G}_{inter},\mathcal{G}_{intra})$ to efficiently mine relations among the cross-modal representation. In each iteration, there are two main steps for the cross-modal correlation capturing: 1. Node message aggregation: both video and query learn attention contexts from each other; 2. Node representation update: after aggregating the contextual information, video/query embedding updates itself to enhance its semantic understanding.

\subsection{Jointly Cross- and Self-modal Graph}
% Regarding video activity localization, besides the video clip contents and the sentence word meaning, their inter-dependencies and temporal correlations play an even more important role. The interdependency means the relation between the frame-words pair, where the more relevant pairs need to be associated with each other. And it is also not enough to only establish the inter-dependencies of frame-word pair one-by-one, as the rich semantic indications of the sentence also provides crucial information to temporally compose the consecutive video contents over time. On the other hand, the temporal correlation of video contexts also help to compose the meaning of the sentence. Based on the above considerations, in this section, we propose a novel inter-intra modality attentive graph network, which first dynamically modulate the feature composition process by considering such semantic information in cross-modal representation, and then further capture the within-modality relations for reconstruction.
% In this stage, we need to locate the moment in video with the help of query.
As shown in Figure \ref{fig:edge}, we develop a jointly cross- and self-modal graph to capture both cross- and self-modal information for multi-modal representation interaction. Specially, the joint graph consists of two subgraphs: cross-modal relation graph (CMG) and self-modal relation graph (SMG). In the CMG, each frame (word) integrates information from the other modality according to the cross-modal attentive relations. Subsequently, in the SMG, the self-attentive contexts within modality are further captured. By stacking $L$ layers of such joint graphs, we can comprehensively perform the interaction between two modalities. Next, we will describe the detailed process of each subgraph in the $l$-th joint graph layer. 

\subsubsection{Cross-Modal Relation Graph}
\ \\
\textbf{Graph construction.} In the CMG, we build a directed graph as $\mathcal{G}_{cross}=(\mathcal{V}_{cross}, \mathcal{E}_{cross})$, where $\mathcal{V}_{cross} = \bm{V} \cup \bm{Q} = \{\bm{v}_t\}_{t=1}^{N_v} \cup \{\bm{q}_n\}_{n=1}^{N_q}$, containing all frames and words as nodes, and $\mathcal{E}_{cross}$ is the edge set between all word-frame node pairs, namely edge $\bm{e}_{(n,t)}=(\bm{q}_n,\bm{v}_t)$ represents the cross-modal interaction from word node $\bm{q}_n$ to frame node $\bm{v}_t$ and $\bm{e}_{(t,n)}=(\bm{v}_t,\bm{q}_n)$ denotes the reverse interaction. 
To initialize the input features for each node,
we set the encoded video representation of nodes $\bm{V}$ and query representation of nodes $\bm{Q}$ as initial hidden states: $\text{H}^{0}(\bm{V})=\hat{\bm{V}}$ and $\text{H}^{0}(\bm{Q})=\hat{\bm{Q}}$ in the CMG, respectively. 
% The initial input features for nodes in the CMG are the encoded video representation $H^{0}[\bm{V}]=\hat{\bm{V}}=\{\hat{\bm{v}}_t\}_{t=1}^{N_v}$ and query representation $H^{0}[\bm{Q}]=\hat{\bm{Q}}=\{\hat{\bm{q}}_n\}_{n=1}^{N_q}$, where $\hat{\bm{v}}_t$ and $\hat{\bm{n}}_q$ are feature vectors for node $\bm{v}_t$ and $\bm{n}_q$, respectively. 
\\ 
% The goal of CMG is to discover the cross-modal interaction between two modalities and can be decomposed into three components: cross-modal attention, node message aggregation
% At $l$-th layer of the jointly graph, CMG contains nodes $\mathcal{V}_{cross} = \hat{\bm{V}}^{l-1} \cup \hat{\bm{Q}}^{l-1}$, where $\hat{\bm{V}}^{l-1}=\{\hat{\bm{v}}_t^{l-1}\}_{t=1}^{N_v}$ and $\hat{\bm{Q}}^{l-1}=\{\hat{\bm{q}}_n^{l-1}\}_{n=1}^{N_q}$ are word/frame nodes from the last graph layer. 
% Each edge $e^l_{(t,n)}=(\hat{\bm{v}}_t^{l-1}, \hat{\bm{q}}_n^{l-1})\in \mathcal{E}_{cross}$ links two modalities nodes $\hat{\bm{v}}_t^{l-1}$ and $\hat{\bm{q}}_n^{l-1}$, and mines the implicitly relation from $\hat{\bm{v}}_t^{l-1}$ to $\hat{\bm{q}}_n^{l-1}$.\\
\textbf{Cross-modal attention.} To update CMG, the first step is to compute the attention weights between frame and word nodes $\bm{V}$, $\bm{Q}$ which represent their pair-wise relations. As in Figure \ref{fig:edge} (a), the attention weight on each pair-wise edge can be computed as below:
\begin{equation}
    \bm{e}^l = (\text{H}^{l-1}(\bm{Q})\cdot\bm{W_q})(\text{H}^{l-1}(\bm{V})\cdot\bm{W_v})^T \in \mathbb{R}^{N_q \times N_v}.
    \label{eq:embedding}
\end{equation}
$\text{H}^{l-1}(\bm{Q})$ and $\text{H}^{l-1}(\bm{V})$ are the feature vectors for word and frame nodes in $(l-1)$-th layer. As $\bm{Q}$ and $\bm{V}$ come from different feature distributions, $\bm{W_q},\bm{W_v} \in \mathbb{R}^{d \times d}$ are linear projection used to embed the heterogeneous nodes \cite{hu2020heterogeneous} into a joint latent space instead of direct computing in the node embedding space. Each row of $\bm{e}^l$ denotes the similarity of all frame nodes $\bm{V}$ to the specific word node ${\bm{q}_n}$, and each column represents the similarity of all word nodes $\bm{Q}$ to the specific frame node ${\bm{v}_t}$.
\\
\textbf{Node message aggregation.} For message aggregation, we aggregate the assigned features for each node from its neighbors in an edge-weighted manner \cite{velivckovic2017graph}. 
% Similarly, each value of the $t$-th column $\bm{e}^l_t = \{\bm{e}^l_{(n,t)}\}^{N_q}_{n=1}$ refers to the semantic correlation of word-wise node $\{\hat{\bm{q}}_n^{l-1}\}$ to frame-wise node $\hat{\bm{v}}_t^{l-1}$. 
% By attending to each node pair, the line-edges can explore their joint representations. 
Here we introduce Figure \ref{fig:edge} (b) which aggregates all neighboring word nodes $\{\bm{q}_n\}_{n=1}^{N_q}$ for frame node ${\bm{v}_t}$, and Figure \ref{fig:edge} (c) performs the reverse aggregation process. For word node $\bm{q}_n$ in $\bm{Q}$, the assigned feature to ${\bm{v}_t}$ is:  
% Given a frame node $\hat{\bm{v}}_t^{l-1}$ whose neighborhood is the entire word nodes $\hat{\bm{Q}}^{l-1}$, the subtle message passing from one neighbor $\hat{\bm{q}}_n^{l-1}$ to the node $\hat{\bm{v}}_t^{l-1}$ is:
\begin{equation}
    \bm{M}^l_{(n,t)} = \text{Softmax}(\bm{e}^l_{(n,t)})\text{H}^{l-1}(\bm{q}_n) \in \mathbb{R}^{1\times d}.
\end{equation}
The softmax procedure makes the sum of all word nodes' attention vectors
to one. 
However, not all neighborhood nodes share same semantic importance, several neighborhood nodes contribute less to target node. For example, word ``the" in the query is not informative enough to the frame, and the frames only containing one stationary basketball should have less significance to highlight ``playing basketball".
To emphasize informative neighborhood nodes and weaken inessential ones, we apply a learnable gate function $G(\cdot)$ to measure the confidence of each neighbor message by:
\begin{equation}
    \bm{g}^l_{(n,t)} = G(\bm{M}^l_{(n,t)}) = \sigma(\bm{M}^l_{(n,t)}\bm{W}_g+\bm{b}_g) \in (0,1),
\end{equation}
where $\sigma(\cdot)$ is the sigmoid function, $\bm{W}_g\in\mathbb{R}^{d\times 1}$ and $\bm{b}_g\in\mathbb{R}^{1}$ are the trainable weight parameter and bias. Then, we can aggregate the gated messages for node ${\bm{v}}_t$ by:
\begin{equation}
    \bm{M}_t^l = \sum_{n=1}^{N_q} \bm{g}^l_{(n,t)}\bm{M}_{(n,t)}^l \in \mathbb{R}^{1 \times d}.
\end{equation}
With the help of such gate mechanism, the irrelevant aggregated messages are filtered and messages from relevant node pairs are further enhanced.\\
\textbf{Node representation update.}
After aggregating the information from all neighbors, node $\bm{v}_t$ gets a new state by taking into account its current state and its received messages $\bm{M}_t^l$. To preserve the sequential information conveyed in the prior state and the messages, we do not utilize a simple element-wise matrix addition on $\text{H}^{l-1}(\bm{v}_t)$ and $\bm{M}_t^l$. Instead, we leverage a ConvGRU \cite{ballas2015delving} layer to update the node state with two inputs by:
\begin{equation}
    \bar{\text{H}}^{l}(\bm{v}_t) = \text{ConvGRU}(\text{H}^{l-1}(\bm{v}_t), \bm{M}_t^l) \in \mathbb{R}^{1 \times d}.
\end{equation}
This ConvGRU is proposed as a convolutional counterpart to original fully connected GRU \cite{cho2014learning}. In the same way, the representations for nodes of two modalities can be updated. 
% denoted as $\hat{\bm{V}}^l_{cross} = \{(\hat{\bm{v}}_t^l)_{cross}\}_{t=1}^{N_v} \in \mathbb{R}^{N_v \times d}$ and $\hat{\bm{Q}}^l_{cross} = \{(\hat{\bm{q}}_n^{l})_{cross}\}_{n=1}^{N_q} \in \mathbb{R}^{N_q \times d}$.

\subsubsection{Self-Modal Relation Graph}
\ \\
\textbf{Graph construction.}
Following CMG, our SMG aims to capture the complex self-modal relations within each modality. It only connects edges between word-word or frame-frame node pairs. Like CMG, we denote this graph as $\mathcal{G}_{self}=(\mathcal{V}_{self}, \mathcal{E}_{self})$, where $\mathcal{V}_{self}$ is the node set containing all frames and words, and each edge in edge set $\mathcal{E}_{inter}$ indicates the self-modal relation. 
% The neighbors $\mathcal{N}_u$ of node $\mathcal{V}_u \in \mathcal{V}_{self}$ only contains the nodes from the within modality.
\\
\textbf{Self-modal attention.}
% After receiving the cross-modality representations $\hat{\bm{V}}^l_{cross}, \hat{\bm{Q}}^l_{cross}$ at $l$-th graph layer, the SMG is further utilized to enhance the within-modality relations. The learned self-relation of each modality is conditioned on the other modality by previously cross-modal interaction graph. It contributes to distinguish and correlate sequential activities along itself, and is also complementary to the cross-modality relations during the unified cross-self graph iteratively updating.
Figure \ref{fig:edge} (d) and (e) depict the process of self-modal information passing. Given a node $\bm{v}_t$ in Figure \ref{fig:edge} (d) for example, we first compute a self-attention matrix to stand for the relations from its neighbor nodes to itself. To better correlate the relevant nodes, in this stage, we consider both the semantic information as well as the temporal position in the sequence of each node. We argue that the temporal index of the node is critical to our localization task as less attention should be given to distant frame (word) nodes, even if they are semantically similar to the current node. Inspired by Transformer’s positional encoding \cite{vaswani2017attention}, we denote the position encoding for each node as:
\begin{equation}
    \text{PE}(\bm{v}_t) = \left\{  
             \begin{array}{cl}
              {\rm sin}(t/10000^{j/d}),  & \text{if}\; j \;  \text{is even}  \\ {\rm cos}(t/10000^{j/d}),  & \text{if}\;  j \;  \text{is odd}  \\
             \end{array},
\right. 
\end{equation}
where $\text{PE}(\bm{v}_t) \in \mathbb{R}^{1 \times d}$, and $j$ varies from 1 to $d$ dimension. With the positional and semantic information combined, the self-attention matrix can be calculated by: 
\begin{equation}
\bar{\bm{e}}^l_t = (\bar{\text{H}}^{l}(\bm{v}_t)+\alpha\text{PE}(\bm{v}_t)) (\bar{\text{H}}^{l}(\bm{V}) + \alpha\text{PE}(\bm{V}))^T,
\end{equation}
where $\bar{\bm{e}}^l_t \in \mathbb{R}^{1 \times N_v}$ calculates the similarity for all frame nodes $\bm{V}$ to current nodes $\bm{v}_t$, $\alpha$ is to balance the two types of information. \\ 
% The positional encodings are helpful to correlate related nodes considering their positions, because the adjacent frames or words generally are more informative. 
% It also can be seen as a gate function to distinguish the closer and remote frames or words.
% Different from CMG, we aggregate messages for node $(\hat{\bm{v}}_t^{l})_{cross}$ in an edge-weighted manner without gating mechanism as each component shares same importance for self-modal relationship exploring.
\textbf{Node message aggregation.}
The aggregation process can be formulated as following:
\begin{equation}
\bar{\bm{M}}_t^l = \text{Softmax}(\bar{\bm{e}}^l_t)\bar{\text{H}}^{l}(\bm{V}) \in \mathbb{R}^{1 \times d}.
\end{equation}
Here we only aggregate the information from $\bar{\text{H}}^{l}(\bm{v}_t)$ as positional information is designed for auxiliary similarity computing.
% and we propose not to introduce additional positional information for later node updating and next layer's CMG. 
% where $\mathcal{V}_u \in \{(\hat{\bm{v}}_t^{l})_{cross}, (\hat{\bm{q}}_n^{l})_{cross}\}$ and $\mathcal{N}_u^l \in \{\hat{\bm{V}}^l_{cross},\hat{\bm{Q}}^l_{cross}\}$.
\\
\textbf{Node representation update.}
Similar to CMG, we also exploit a ConvGRU layer to update the node state and get the output as:
\begin{equation}
    \text{H}^{l}(\bm{v}_t) = \text{ConvGRU}(\bar{\text{H}}^{l}(\bm{v}_t),\bar{\bm{M}}_t^l) \in \mathbb{R}^{1 \times d}.
\end{equation}
At last, following the same procedure, we can get the final representations for all nodes of each modality in the $l$-th jointly cross- and self-modal graph layer as $\text{H}^{l}(\bm{V}) \in \mathbb{R}^{N_v \times d}$ and $\text{H}^{l}(\bm{Q}) \in \mathbb{R}^{N_q \times d}$. Subsequently, these two modal features will be feed to the $(l+1)$-th joint graph layer as input.

\subsection{Multi-Modal Integration Module}
After $L$ joint graph layer, we can get mutual sentence-aware video representation $\tilde{\bm{V}}=\{\tilde{\bm{v}}_t\}_{t=1}^{N_v}=\text{H}^{L}(\bm{V})$ and video-aware sentence representation $\tilde{\bm{Q}}=\{\tilde{\bm{q}}_n\}_{n=1}^{N_q}=\text{H}^{L}(\bm{Q})$. To integrate both two representations, 
% we need to project $\tilde{\bm{Q}}$ into the same dimension like $\tilde{\bm{V}}$ for concatenation. 
we first compute the cosine similarity between each pair of word feature $\tilde{\bm{q}}_n$ and frame feature $\tilde{\bm{v}}_t$ as:
\begin{equation}
    c_{n,t} = \frac{(\tilde{\bm{v}}_t)(\tilde{\bm{q}}_n\bm{W}_c)^T}{\parallel \tilde{\bm{v}}_t \parallel_2 \parallel \tilde{\bm{q}}_n\bm{W}_c \parallel_2},
\end{equation}
% where $\bm{W}_c \in \mathbb{R}^{d \times d}$ is the parameter to transform word into the same dimensional space as the video frame features. This similarity scores are non-directional, and different to the relation weights in our CMG. It is more suitable here to fuse the query clues for each frame.  
where $\bm{W}_c \in \mathbb{R}^{d \times d}$ is the linear parameter. $c_{n,t}$ is used to further extract the implicit query clues for each frame.
Once we get the similarity scores between entire words $\{\tilde{\bm{q}}_n\}_{n=1}^{N_q}$ and a specific frame $\tilde{\bm{v}}_t$, 
% we apply a softmax operation along the temporal dimension of $\bm{c_t}=\{c_{t,n}\}^{N_q}_{n=1}$ to obtain an attention vector for extracting the crucial query clues as:
we integrate the query information for frame $\tilde{\bm{v}}_t$ as:
\begin{equation}
    \bm{h}_t = \sum_{n=1}^{N_q} \text{Softmax}(c_{n,t}) \tilde{\bm{q}}_n \in \mathbb{R}^{1 \times d},
\end{equation}
where $\bm{h}_t$ is aggregated with the query representation relevant to the $t$-th frame. We concat such aggregated query feature vectors with frame features to get the final multi-modal semantic representations $\bm{f} = \{\bm{f}_t\}_{t=1}^{N_v}$, where $\bm{f}_t = \text{Concat}[\tilde{\bm{v}}_t,\bm{h}_t] \in \mathbb{R}^{1 \times 2d}$.

\subsection{Moment Localization}
We first apply a bi-directional GRU network on $\bm{f}$ to further absorb the contextual evidences in temporal domain. To predict the target video segment, we pre-define a set of candidate moments $\Phi_t = \{(\hat{s}_{t, i}, \hat{e}_{t, i})\}_{i=1}^{N_{\Phi}}$ with multi-scale windows \cite{yuan2019semantic} at each time $t$, where $N_{\Phi}$ is the number of moments at current time-step. Then, we need to score these candidate moments and predict the offsets $\hat{\delta}_t = \{(\hat{\delta}_{t,i}^s, \hat{\delta}_{t,i}^e)\}_{i=1}^{N_{\Phi}}$ of them relative to the ground-truth. In details, we produce the confidence scores $cs_t = \{cs_{t,i}\}_{i=1}^{N_{\Phi}}$ for these moments at time $t$ by a Conv1d layer:
\begin{equation}
    cs_{t,i} = \sigma(\text{Conv1d}(\bm{f}_t)) \in (0,1),
\end{equation}
where $\sigma(\cdot)$ is the sigmoid function. The temporal offsets are predicted by another Conv1d layer: 
\begin{equation}
    (\hat{\delta}_{t,i}^s, \hat{\delta}_{t,i}^e) = \text{Conv1d}(\bm{f}_t).
\end{equation}
Therefore, the final predicted moment $i$ of time $t$ can be presented as $(\hat{s}_{t,i}+\hat{\delta}_{t,i}^s, \hat{e}_{t, i}+\hat{\delta}_{t,i}^e)$.
\\
\textbf{Training.}
We first compute the IoU (Intersection over Union) score $IoU_{t,i}$ between each candidate moment $(\hat{s}_{t,i}, \hat{e}_{t, i})$ with the ground truth $(s_t, e_t)$. If $IoU_{t,i}$ is larger than an IoU threshold $\tau$, we treat this candidate moment as a positive sample.
We adopt an alignment loss to learn the confidence scoring rule for candidate moments, where the moments with higher IoUs will get higher confidence scores. The alignment loss function can be formulated as follows:
\begin{equation}
    \mathcal{L}_{align} = - \frac{1}{N_vN_{\Phi}} \sum_{t=1}^{N_v}\sum_{i=1}^{N_{\Phi}} IoU_{t,i} log(cs_{t,i}) + (1-IoU_{t,i}) log(1-cs_{t,i}).
\end{equation}
Since parts of the pre-defined candidates are coarse in boundaries, we only fine-tune the localization offsets of positive moment samples by a boundary loss:
\begin{equation}
    \mathcal{L}_{b} = \frac{1}{N_{pos}} \sum_j^{N_{pos}} \mathcal{R}_1(\hat{\delta}_j^s-\delta_j^s) + \mathcal{R}_1(\hat{\delta}_j^e-\delta_j^e),
\end{equation}
where $N_{pos}$ denotes the number of positive moments, and $\mathcal{R}_1$ is the smooth L1 loss. Therefore, the joint loss can be represented as:
\begin{equation}
    \mathcal{L} = \mathcal{L}_{align} + \beta \mathcal{L}_b,
\end{equation}
where $\beta$ is utilized to control the balance.
\\
\textbf{Inference.}
We first rank all candidate moments according to their predicted confidence scores, and then adopt a non-maximum suppression (NMS) to select ``Top $n$" moments as the prediction.

\section{EXPERIMENTS}
\subsection{Datasets and Evaluation Metrics}
% We conduct experiments on Activity Caption \cite{krishna2017dense}, TACoS \cite{regneri2013grounding}, Charades-STA \cite{gao2017tall}, and DiDeMo \cite{anne2017localizing} datasets. 
% Due to the limited space of the paper, we present dataset descriptions, experimental settings, and performance results on Charades-STA and DiDeMo datasets in the supplementary material.

\noindent\textbf{Activity Caption.}
Activity Caption \cite{krishna2017dense} contains 20k untrimmed videos with 100k descriptions from YouTube. The videos are 2 minutes on average, and the annotated video clips have much larger variation, ranging from several seconds to over 3 minutes. Since the test split is withheld for competition, following public split, we adopt ``val 1” as validation subset, ``val 2” as our test subset.

\noindent \textbf{TACoS.}
TACoS \cite{regneri2013grounding} is widely used on this task and contain 127 videos. The videos from TACoS are collected from cooking scenarios, thus lacking the diversity. They are around 7 minutes on average. We use the same split as \cite{gao2017tall}, which includes 10146, 4589, 4083 query-segment pairs for training, validation and testing.

\noindent \textbf{Evaluation Metrics.}
Following previous works \cite{gao2017tall,yuan2019semantic}, we adopt ``R@n, IoU=m” as our evaluation metrics. The “R@n, IoU=m” is defined as the percentage of at least one of top-n selected moments having IoU larger than m. 
Following \cite{liu2018attentive,yuan2019semantic,wang2019temporally}, we choose the evaluation criteria “R@n, IoU=m” with $\text{n} \in \{1, 5\}, \text{m} \in \{0.3, 0.5, 0.7\}$ and “R@n, IoU=m” with $\text{n} \in \{1, 5\}, \text{m} \in \{0.1, 0.3, 0.5\}$ for Activity Caption and TACoS datasets, respectively.

\begin{table}[t!]
    \small
    \centering
    \caption{Performance compared with previous methods on the Activity Caption dataset.}
    \label{tab:compare1}
    \setlength{\tabcolsep}{1.5mm}{
    \begin{tabular}{c|cccccc}
    \hline \hline
    \multirow{2}*{Method} & R@1 & R@1 & R@1 & R@5 & R@5 & R@5 \\ 
    ~ & IoU=0.3 & IoU=0.5 & IoU=0.7 & IoU=0.3 & IoU=0.5 & IoU=0.7 \\ \hline
    MCN \cite{anne2017localizing} & 39.35 & 21.36 & 6.43 & 68.12 & 53.23 & 29.70 \\ 
    % VSA-RNN & 39.28 & 23.43 & 9.01 & 70.84 & 55.52 & 32.12 \\
    % VSA-STV & 41.71 & 24.01 & 8.92 & 71.05 & 56.62 & 34.52 \\
    TGN \cite{chen2018temporally} & 45.51 & 28.47 & - & 57.32 & 43.33 & - \\
    CTRL  \cite{gao2017tall} & 47.43 & 29.01 & 10.34 & 75.32 & 59.17 & 37.54 \\
    ACRN \cite{liu2018attentive} & 49.70 & 31.67 & 11.25 & 76.50 & 60.34 & 38.57 \\
    QSPN \cite{xu2019multilevel} & 52.13 & 33.26 & 13.43 & 77.72 & 62.39 & 40.78 \\
    CBP \cite{wang2019temporally} & 54.30 & 35.76 & 17.80 & 77.63 & 65.89 & 46.20 \\
    SCDM \cite{yuan2019semantic} & 54.80 & 36.75 & 19.86 & 77.29 & 64.99 & 41.53 \\
    ABLR \cite{yuan2019find} & 55.67 & 36.79 & - & - & - & - \\
    GDP \cite{chenrethinking} & 56.17 & 39.27 & - & - & - & - \\
    CMIN \cite{zhang2019cross} & 63.61 & 43.40 & 23.88 & 80.54 & 67.95 & 50.73 \\ \hline
    \textbf{CSMGAN} & \textbf{68.52} & \textbf{49.11} & \textbf{29.15} & \textbf{87.68} & \textbf{77.43} & \textbf{59.63} \\ \hline
    \end{tabular}}
\end{table}

\begin{table}[t!]
    \small
    \centering
    \caption{Performance compared with previous methods on the TACoS dataset.}
    \label{tab:compare2}
    \setlength{\tabcolsep}{1.5mm}{
    \begin{tabular}{c|cccccc}
    \hline \hline
    \multirow{2}*{Method} & R@1 & R@1 & R@1 & R@5 & R@5 & R@5 \\ 
    ~ & IoU=0.1 & IoU=0.3 & IoU=0.5 & IoU=0.1 & IoU=0.3 & IoU=0.5 \\ \hline
    MCN \cite{anne2017localizing} & 3.11 & 1.64 & 1.25 & 3.11 & 2.03 & 1.25 \\ 
    % VSA-RNN & 8.84 & 10.77 & 4.78 & 19.05 & 13.90 & 9.10 \\
    % VSA-STV & 15.01 & 10.77 & 7.56 & 32.82 & 23.92 & 15.50 \\
    CTRL  \cite{gao2017tall} & 24.32 & 18.32 & 13.30 & 48.73 & 36.69 & 25.42 \\
    ABLR \cite{yuan2019find} & 34.70 & 19.50 & 9.40 & - & - & - \\
    ACRN \cite{liu2018attentive} & 24.22 & 19.52 & 14.62 & 47.42 & 34.97 & 24.88 \\
    QSPN \cite{xu2019multilevel} & 25.31 & 20.15 & 15.23 & 53.21 & 36.72 & 25.30 \\
    TGN \cite{chen2018temporally} & 41.87 & 21.77 & 18.90 & 53.40 & 39.06 & 31.02 \\
    GDP \cite{chenrethinking} & 39.68 & 24.14 & 13.50 & - & - & - \\
    CMIN \cite{zhang2019cross} & 32.48 & 24.64 & 18.05 & 62.13 & 38.46 & 27.02 \\
    SCDM \cite{yuan2019semantic} & - & 26.11 & 21.17 & - & 40.16 & 32.18 \\
    CBP \cite{wang2019temporally} & - & 27.31 & 24.79 & - & 43.64 & 37.40 \\
    \hline
    \textbf{CSMGAN} & \textbf{42.74} & \textbf{33.90} & \textbf{27.09} & \textbf{68.97} & \textbf{53.98} & \textbf{41.22} \\ \hline
    \end{tabular}}
    \vspace{-5pt}
\end{table}

\begin{table*}[t!]
\small
\centering
\caption{Ablation study on the Activity Caption and TACoS datasets, where the reference is our full model.}
\label{tab:ablation12}
\begin{tabular}{c|c|cccccc|cccccc}
\hline
\multirow{3}*{Components} & \multirow{3}*{Module} &  \multicolumn{6}{c|}{Activity Caption} & \multicolumn{6}{c}{TACoS} \\ \cline{3-8} \cline{9-14}
~ & ~ & R@1 & R@1 & R@1 & R@5 & R@5 & R@5 & R@1 & R@1 & R@1 & R@5 & R@5 & R@5 \\ 
~ & ~ & IoU=0.3 & IoU=0.5 & IoU=0.7 & IoU=0.3 & IoU=0.5 & IoU=0.7 & IoU=0.1 & IoU=0.3 & IoU=0.5 & IoU=0.1 & IoU=0.3 & IoU=0.5 \\ \hline \hline
Reference & \textbf{full} & \textbf{68.52} & \textbf{49.11} & \textbf{29.15} & \textbf{87.68} & \textbf{77.43} & \textbf{59.63} & \textbf{42.74} & \textbf{33.90} & \textbf{27.09} & \textbf{68.97} & \textbf{53.98} & \textbf{41.22} \\ \hline \hline
Encoder & \textbf{w/o HS} & 66.32 & 46.80 & 26.54 & 85.97 & 74.43 & 56.49 & 39.56 & 30.42 & 24.40 & 65.50 & 51.39 & 39.38 \\ \hline
Joint Graph & \textbf{w/o CSG} & 64.13 & 44.47 & 25.49 & 84.35 & 72.97 & 54.47 & 36.91 & 28.45 & 22.27 & 63.18 & 49.19 & 37.11 \\ \hline
\multirow{2}*{\tabincell{c}{Cross-Modal\\ Graph}} & \textbf{w/o EM} & 67.48 & 47.94 & 28.09 & 86.02 & 75.27 & 56.32 & 41.11 & 32.24 & 25.93 & 66.67 & 53.05 & 40.17 \\
~ & \textbf{w/o MG} & 67.28 & 47.39 & 28.13 & 86.46 & 74.91 & 56.47 & 40.48 & 31.66 & 25.73 & 66.39 & 52.56 & 40.21 \\ \hline
\multirow{2}*{\tabincell{c}{Self-Modal\\ Graph}} & \textbf{w/o SMG} & 66.53 & 46.62 & 27.57 & 85.68 & 73.97 & 55.68 & 39.64 & 30.86 & 24.61 & 65.10 & 50.90 & 39.11 \\
~ & \textbf{w/o PE} & 67.41 & 48.45 & 28.56 & 86.51 & 75.20 & 57.20 & 40.23 & 31.32 & 25.30 & 66.17 & 51.41 & 39.43 \\ \hline
Node Update& \textbf{w/o CG} & 67.37 & 47.51 & 28.07 & 86.06 & 75.66 & 56.96 & 40.97 & 31.96 & 25.66 & 66.42 & 52.00 & 40.04 \\ 
\hline
% Integration & \textbf{w/o MI} & 66.18 & 47.14 & 27.24 & 87.09 & 75.04 & 56.42 & 36.32 & 25.06 & 20.28 & 64.23 & 48.17 & 33.60 \\ \hline
\end{tabular}
\vspace{-8pt}
\end{table*}

\subsection{Implementation Details}
For training our CSMGAN, we first resize every frame of videos to $112 \times 112$ pixels as input, and then apply a pre-trained C3D \cite{tran2015learning} to obtain 4096 dimension features. After that we apply PCA to reduce the feature dimension from 4096 to 500 for decreasing the model parameters. These 500-d features are used as the frame features in our model. Since some videos are overlong, we set the length of video feature sequences to 200 for both Activity Caption and TACoS datasets. As for sentence encoding, we utilize Glove word2vec \cite{pennington2014glove} to embed each word to 300 dimension features. The hidden state dimension of BiGRU networks is set to 512. We set $\alpha$ to 1 for positional encoding. During moment localization, we adopt convolution kernel size of [16, 32, 64, 96, 128, 160, 192] for Activity Caption, and [8, 16, 32, 64] for TACoS. We set the stride of them as 0.5, 0.125, respectively. We then set the high-score threshold $\tau$ to 0.45, and the balance hyper-parameter $\beta$ to 0.001 for Activity Caption, 0.005 for TACoS. The number of our joint graph layer is set to 2. We train our model with an Adam optimizer with leaning rate $8\times 10^{-4}, 3\times 10^{-4}$ for Activity Caption and TACoS, respectively. The batch size is set to 128 and 64 for two datasets, respectively.

\begin{figure}
    \centering
    \subfloat{\includegraphics[width=0.23\textwidth]{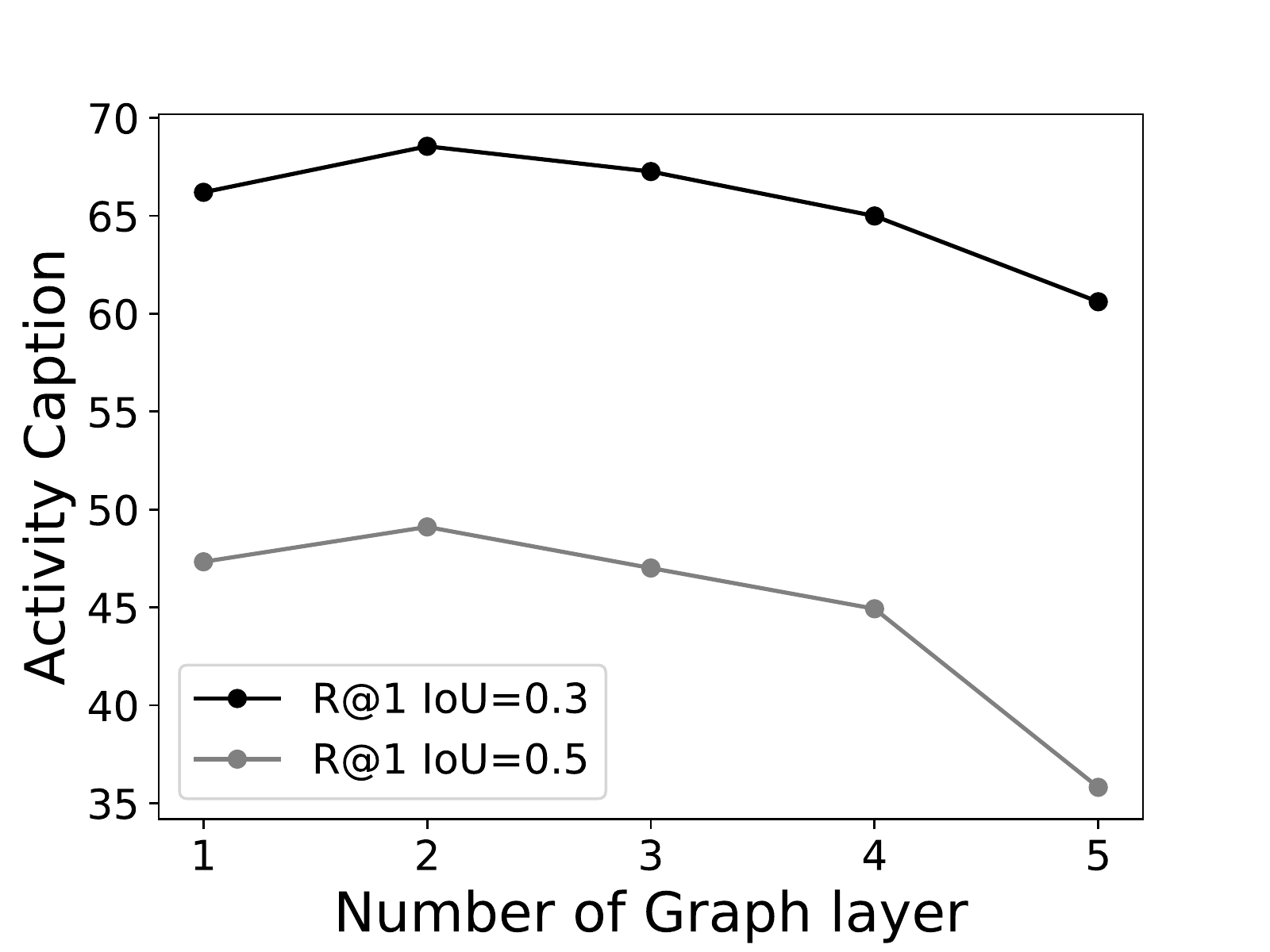}}
    \subfloat{\includegraphics[width=0.23\textwidth]{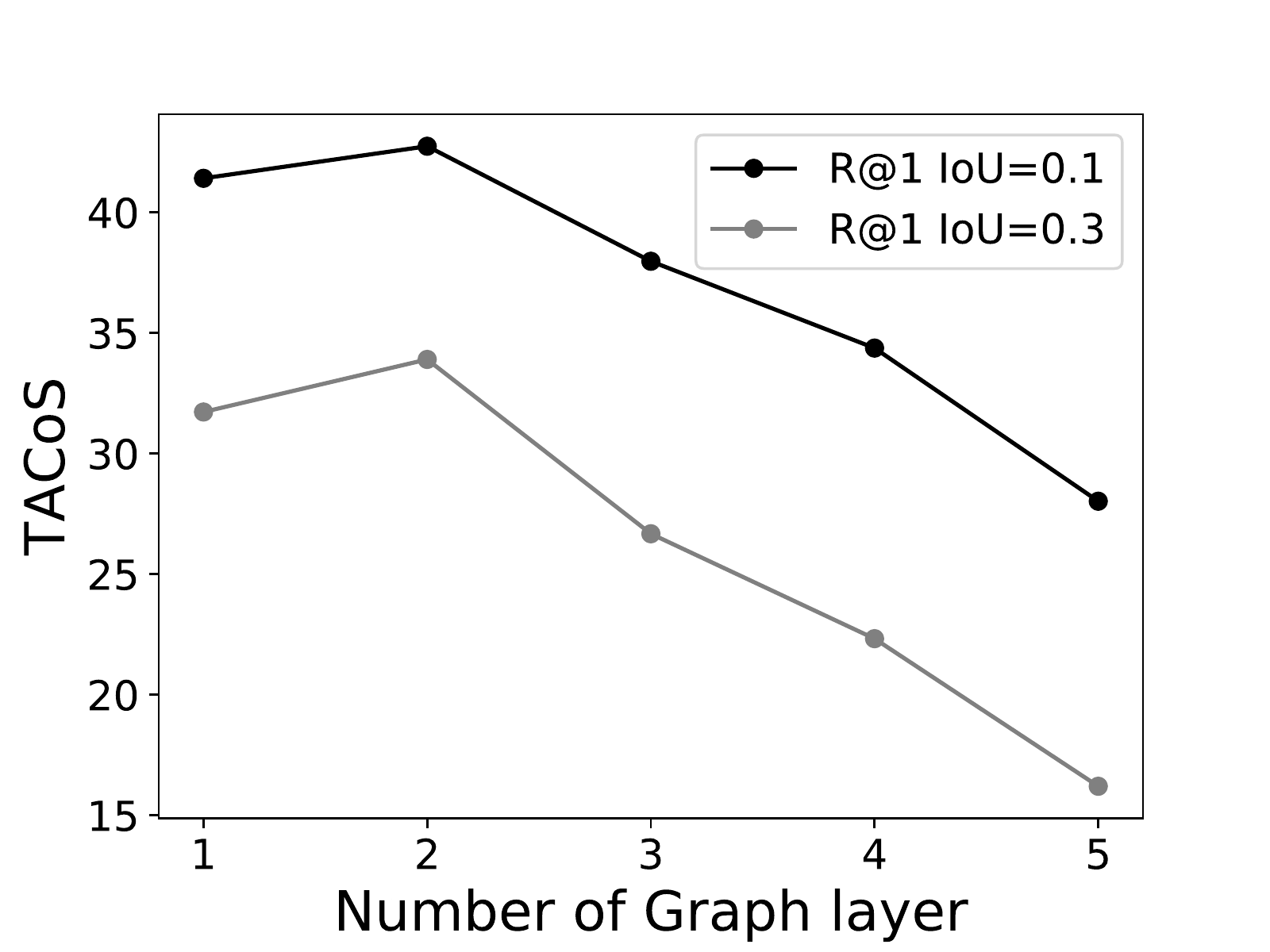}}
    \caption{Effect of the number of graph layers on the Activity Caption and TACoS Datasets.}
    \label{fig:graph_layer}
    \vspace{-12pt}
\end{figure}

\subsection{Performance Comparison and Analysis}
\textbf{Activity Caption.}
Table \ref{tab:compare1} shows the performance evaluation results of our method and all comparing methods on Activity Caption dataset. Compared to the state-of-the-arts methods, our model surpasses them with clear margin on all R@1 and R@5 metrics. Specially, our method brings 5.27\% and 8.90\% absolute improvements in the strict metrics ``R@1, IoU=0.7”  and ``R@5, IoU=0.7”.

\noindent \textbf{TACoS.}
Table \ref{tab:compare2} shows the performance results of our method and all baselines on TACoS dataset. 
% Different from Activity Caption, TACoS is a fine-grained dataset limited to cooking scenes, the videos of TACoS are longer than Activity Caption but the target moments are shorter, which makes this dataset more difficult. Therefore, the overall experimental results on TACoS are lower than Activity Caption. 
On this challenging dataset, we find that our method still achieves significant improvements. In details, our method brings 2.30\% and 3.82\% improvements in the strict metrics``R@1, IoU=0.5”  and ``R@5, IoU=0.5”, respectively.

\noindent \textbf{Analysis.}
Specifically, the compared methods can be divided into two classes: 1) Sliding window based methods: MCN \cite{anne2017localizing}, CTRL \cite{gao2017tall}, and ACRN \cite{liu2018attentive} first sample candidate video segments using sliding windows, and directly integrate query representations with window-based segment representations via a matrix operation. They do not employ a comprehensively structure for effective cross-modal interaction, leading to relatively lower performances than other methods. 2) Cross-modal interaction based methods: TGN \cite{chen2018temporally}, QSPN \cite{xu2019multilevel}, CBP \cite{wang2019temporally}, SCDM \cite{yuan2019semantic}, ABLR \cite{yuan2019find}, GDP \cite{chenrethinking}, and CMIN \cite{zhang2019cross} integrate query representations with the whole video representations in an attention-guided manner, and can generate contextual query-guided video representation for precisely boundary localization. However, they ignore to capture the self-modal relation which helps to correlate relevant instances within each modality. 
Compared to them, our method emphasizes the importance of capturing both cross- and self-modal relation during the effective integration of multi-modal feature. Our jointly cross- and self-modal graph can mine much richer and higher-level interactions, thus achieving better results than both two kinds of methods.

\subsection{Ablation Study}
In this section, we perform ablation studies to examine the effectiveness of our proposed CSMGAN. Specifically, we re-train our model with the following settings:
\begin{itemize}
    \item \textbf{w/o HS}: We first remove the hierarchical structure from the sentence encoder, and only take a bi-directional GRU to encode the sentence query.
    \item \textbf{w/o CSG}: We then discard the jointly cross- and self-modal graph to validate the importance of cross- and self-modal relations capturing in multi-modal interaction.
    \item \textbf{w/o EM}: To explore the effect of heterogeneous attention in CMG, we remove the embedding matrices in Eq. \ref{eq:embedding}, and compute the attentive matrix in the node embedding space.
    \item \textbf{w/o MG}: To further analyze the gate mechanism in CMG, we remove the gate function during the message passing in the cross-modal graph layer.
    \item \textbf{w/o SMG}: To evaluate the effect of SMG, we remove the self-modal graph, and only apply cross-modal graph for multi-modal interaction.
    \item \textbf{w/o PE}: To assess the component of SMG, we remove the positional encoding from the SMG.
    \item \textbf{w/o CG}: Finally, we replace the ConvGRU with a simple matrix element-wise addition during the node updating.
    % \item \textbf{w/o MI}: We remove the multi-modal integration module, and directly use the video representation for localization.
    \item \textbf{full}: The full model.
\end{itemize}

\begin{figure*}
    \centering
    \includegraphics[width=1.0\textwidth]{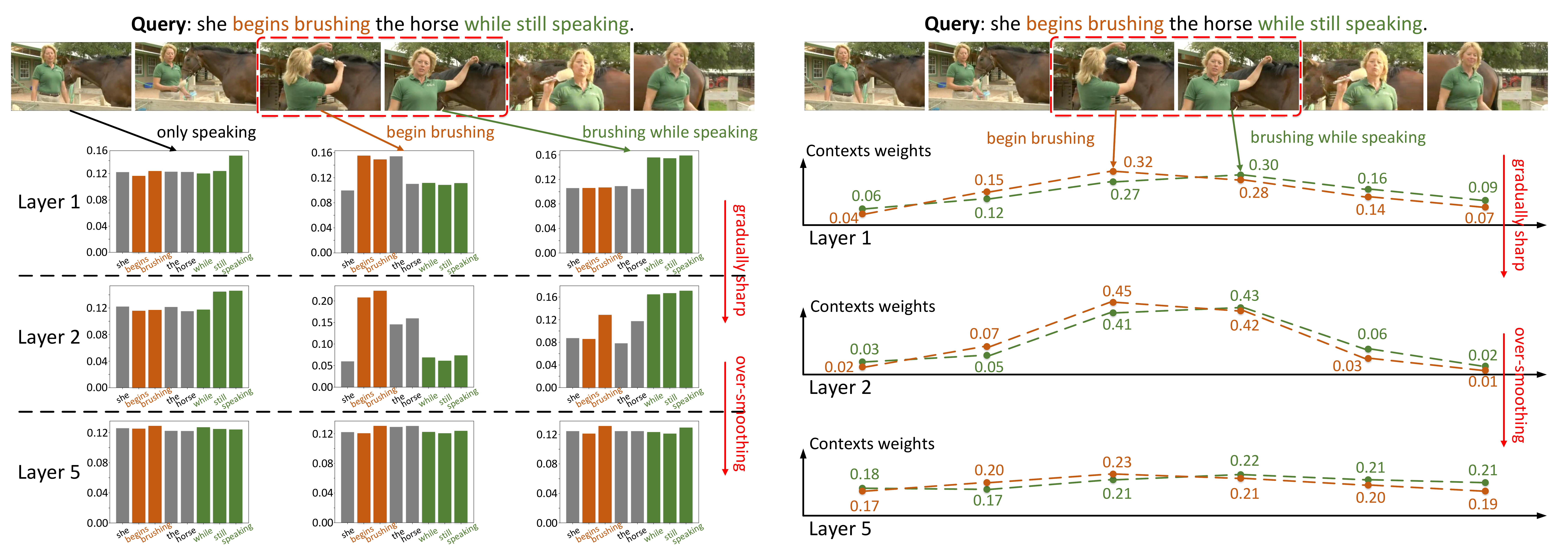}
    \caption{Visualization on the edge weights of both cross-modal graph (CMG) and self-modal graph (SMG). Left: column-wise weights of the attention matrix of different CMG layer, where the weights stands for the relations from all words to a specific frame. Right: self-attention weights of different SMG layer, where the relevant frames have higher context weights.}
    \label{fig:weight}
    \vspace{-8pt}
\end{figure*}

The ablation study conducted on Activity Caption and TACoS datasets are shown in Table \ref{tab:ablation12}. By analyzing the ablation results, we have the conclusions as follows:
\begin{itemize}
    \item First of all, our full model outperforms all the ablation models on both two datasets, which demonstrates each component is definitely helpful for this task.
    \item Compared to other ablation models, the w/o CSG model performs worst on two datasets. This means that our jointly cross- and self-modal graph takes an important role in effective multi-modal features interaction. Besides, the hierarchical structure (HS) for sentence embedding also has significant contribution to the full model.
    % \item Each component of CSMGAN contributes more on TACoS. It means that our method performs well on challenging dataset.
    % \item Specially, in our cross-modal graph, the self-modal graph (SMG), embedding matrices (EM), message gate (MS), and  ConvGRU (CG) all have contributions for the final results. 
    % The SMG conditioned on the other modal features is complementary to the cross-modal relations. The EM mines deeper correlation between the two modalities. The message gate helps to filter out the noise words/frames for each frame/word. And the ConvGRU can preserve the sequential information rather than the element-wise addition. 
    \item At last, almost all ablation models still yield better results than all state-of-the-arts methods. This fact demonstrates that the excellent performance of our graph based framework does not only rely on one specific key component, and our full model is robust to address this task.
\end{itemize}

To further investigate the influence on the various number of our joint graph layers, we show the impact of different layer numbers on two datasets in Figure \ref{fig:graph_layer}. We can observe that our model achieves best result when the number of layer is set to 2. Then the performance will drop if the number of layers increases. The propagated messages between the instances in cross-modality and self-modality will be accumulated if we use more graph layers, resulting in over-smoothing \cite{li2018deeper} problem, namely the representations of both video and sentence converge to the same value.

% \begin{table}
%     \small
%     \centering
%     \caption{Comparison of model running efficiency, model size, and gpu memory.}
%     \label{tab:time}
%     \begin{tabular}{c|ccc}
%     \hline \hline
%     Method & Time/video & Model Size & GPU Memory \\ \hline
%     CTRL & 2.23s & 22M & 725M \\ 
%     ACRN & 4.31s & 128M & 8537M \\
%     SCDM & 0.78s & 15M & 4533M \\ \hline
%     Ours & 0.05s & 68M & 963M \\ \hline
%     \end{tabular}
% \end{table}

% \subsection{Model Efficiency Comparison}
% We also investigate the run-time efficiency, model
% size (\#param), and memory footprint of different methods in Table \ref{tab:time}. “Time/video” means the average time to localize one sentence in a given video. The methods with released codes are run on TACoS dataset with one Nvidia TITAN XP GPU. From the table, we find that our method only take 0.05s to inference a video, which achieves the fastest time compared to others. Also, our model has relatively small model size and gpu memory. The CTRL and ACRN methods are quite time-consuming by the matching procedure through various sliding window. And SCDM adopts a hierarchical convolution architecture which serves a large memory consuming. Different from them, our method covers multi-scale video segments for grounding and only need to inference one pass for each video. And our model has no many convolution operation. Therefore, we can achieve much faster performance with smaller gpu memory consuming.

\begin{figure}
   \centering
    \includegraphics[width=0.45\textwidth]{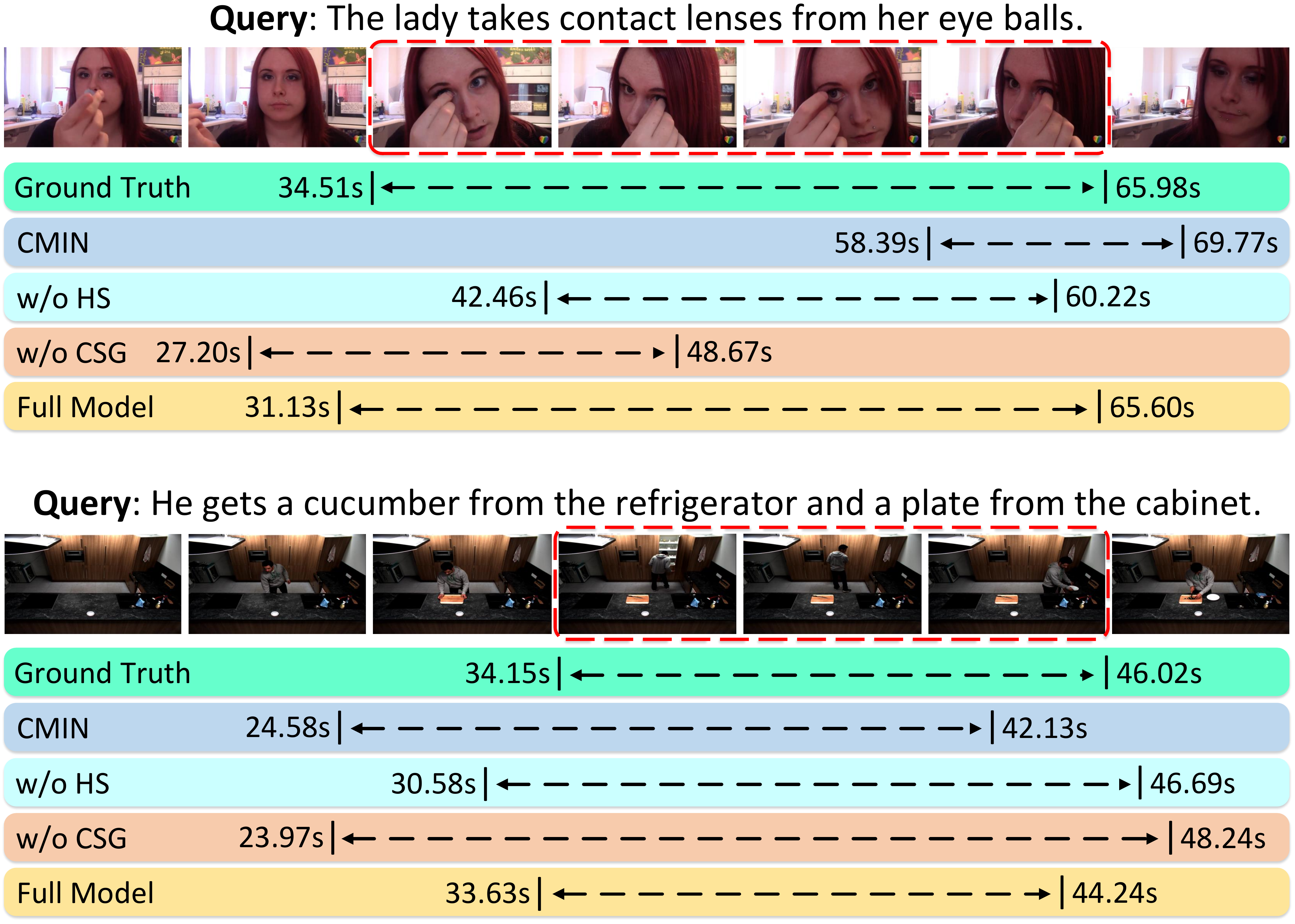}
    \caption{Qualitative visualization on both two datasets (top: Activity Caption, bottom: TACoS).}
    \label{fig:result}
    \vspace{-12pt}
\end{figure}

\subsection{Qualitative Results}
To qualitatively validate the effectiveness of our method, we show examples from two datasets in Figure \ref{fig:result}. Although the sentences are very diverse, our full model can still localize more accurate boundaries than CMIN. In two variant models, the w/o CSG has the most coarse boundaries because it lacks the detailed interaction of multi-modal features. The w/o HS fails to capture more contextual sentence guiding clues for localization, leading to relatively coarse boundaries. As a comparison, our full model achieves the most precise localization.

We further give a deep visualization on the cross- and self-modal relations in each joint graph layer. Specially, we first visualize the relations from all words to a specific frame in the cross-modal graph. As shown in Figure \ref{fig:weight} (left), sentence ``she begins brushing the horse while still speaking" has two activities. For the non-relevant frame, the attention weights on these eight words are more inclined to be an even distribution. But for the relevant frame, the contributed words like ``begins", ``brushing", ``still", and ``speaking" obtain higher attention weights since the described action indeed happens there. Moreover, with GNN layer increasing, the distribution of these words weights are sharper and more distinguishable. However, too many GNN layers will result in over-smoothing problem, where each frame-word pair has almost the same activation. We also plot the context weights in the self-modal graph as shown in Figure \ref{fig:weight} (right). The weights are calculated by a softmax function, and they represent the relations from surrounding frames to one specific frame. We find that frame containing ``brushing while speaking" is more relevant to the frame ``begin brushing". Although the frames near the segment boundaries are visually similar to the frames in the segment, the self-modal relation can effectively distinguish them and produce lower attention weights for such noisy frames.
% \begin{figure}
%     \centering
%     \includegraphics[width=0.5\textwidth]{weight.pdf}
%     \caption{Column-wise weights of the affinity matrix of different CMG layer, where the weights stands for the directional relations from words to the same frame. We can find that more than 2 GNN layers will result in over-smoothing.}
%     \label{fig:weight}
% \end{figure}

% \begin{figure}
%     \centering
%     \includegraphics[width=0.5\textwidth]{weight_other.pdf}
%     \caption{Self-attentive weights of different SMG layer, where the relevant frames of the same activity have higher context weights.}
%     \label{fig:weight_other}
% \end{figure}

% \vspace{-12pt}
\section{Conclusion}
In this paper, we propose a jointly cross- and self-modal graph attention network (CSMGAN) for query-based moment localization in video. We consider both cross- and self-modal relations in a joint framework to capture much higher-level interactions. Specially, 
% the self-modal relations are dynamically conditioned on the other modality, and are complementary to the cross-modal relations. 
cross-modal relation highlights relevant components across video and sentence, and
then self-modal relation models the pairwise correlation inside each modality for
frames/words association. 
Besides, we also develop a hierarchical structure for more contextual sentence understanding in a word-phrase-sentence process. The experimental results on various datasets demonstrate the effectiveness of our proposed method.

\bibliographystyle{ACM-Reference-Format}
\bibliography{sample-base}

\clearpage
% \section{Appendices}
\appendix

\section{Additional Datasets}
\label{sec:analyzes}
In this section, we provide detailed performance analysis on additional datasets, including Charades-STA \cite{gao2017tall} and DiDeMo \cite{anne2017localizing}.

\subsection{Charades-STA}
\noindent \textbf{Dataset.} 
Charades-STA is built on the Charades dataset \cite{sigurdsson2016hollywood}, which focuses on indoor activities. As Charades dataset only provides video-level paragraph description, the temporal annotations of Charades-STA are generated in a semi-automatic way, which involves sentence decomposition, keyword matching, and human check. In total, the video length on the Charades-STA dataset is 30 seconds on average, and there are 12408 and 3720 moment-query pairs in the training and testing sets respectively.

\noindent \textbf{Settings.} Following previous settings \cite{yuan2019semantic,wang2019temporally}, we extract 1024 dimension features by I3D \cite{carreira2017quo}, and then apply PCA to reduce the feature dimension to 500 for decreasing the model parameters. During moment localization, we adopt convolution kernel size of $[16, 24, 32, 40]$, and set the stride as 0.25. We set the high-score threshold $\tau$ to 0.5, and the balance hyper-parameter $\beta$ to 0.005.

\noindent \textbf{Analysis.}
Table \ref{tab:charades} shows the performance evaluation results of our CSMGAN and all comparing methods on Charades-STA dataset. Compared to the state-of-the-arts methods, our model surpasses them with clear margin both on R@1 and R@5 metrics. Specially, compared with the previous state-of-the-art method in absolute values, our method brings 3.91\% and 3.77\% improvements in the strict “R@1, IoU=0.7” and “R@5, IoU=0.7” metrics, respectively.

\subsection{DiDeMo}
\noindent \textbf{Dataset.} 
DiDeMo is recently proposed in \cite{anne2017localizing}, specially for natural language moment retrieval in open-world videos. DiDeMo contains 10464 videos with 33005, 4180 and 4021 annotated moment-query pairs in the training, validation and testing sets respectively. To annotate moment-query pairs, videos in DiDeMo are trimmed to a maximum of 30 seconds, divided into 6 segments of 5 seconds long each, and each moment contains one or more consecutive segments. Therefore, there are 21 candidate moments in each video and the task is to select the moment that best matches the query.

\noindent \textbf{Settings.} Following previous settings \cite{liu2018attentive}, we extract 4096 dimension features by C3D \cite{tran2015learning}, and then apply PCA to reduce the feature dimension to 500 for decreasing the model parameters. During moment localization, we adopt convolution kernel size of $[16, 32, 64, 96]$, and set the stride as 0.25. We set the high-score threshold $\tau$ to 0.5, and the balance hyper-parameter $\beta$ to 0.005.

\noindent \textbf{Analysis.}
Table \ref{tab:didemo} shows the performance comparisons on the DiDeMo dataset. Compared to the state-of-the-arts methods, our model surpasses them with clear margin both on R@1 and R@5 metrics. Specially, compared to the previous state-of-the-art method in absolute values, our method brings 2.51\% improvements in the strict “R@1, IoU=0.7”. At the same time, it is worth noticing that our CSMGAN achieves significant improvements (12.16\%) in the “R@5, IoU=0.7” metrics.

\begin{table}
    \small
    \centering
    \caption{Performance compared with previous methods on the Charades-STA dataset.}
    \label{tab:charades}
    \setlength{\tabcolsep}{1.5mm}{
    \begin{tabular}{c|cccc}
    \hline \hline
    \multirow{2}*{Method} & R@1 & R@1 & R@5 & R@5 \\ 
    ~ & IoU=0.5 & IoU=0.7 & IoU=0.5 & IoU=0.7 \\ \hline
    TGA \cite{mithun2019weakly} & 17.04 & 6.93 & 58.17 & 26.80 \\
    MCN \cite{anne2017localizing} & 17.46 & 8.01 & 48.22 & 26.73 \\ 
    ACRN \cite{liu2018attentive} & 20.26 & 7.64 & 71.99 & 27.79 \\
    CTRL  \cite{gao2017tall} & 23.63 & 8.89 & 58.92 & 29.57  \\
    SAP \cite{chen2019semantic} & 27.42 & 13.36 & 66.37 & 38.15 \\
    QSPN \cite{xu2019multilevel} & 35.60 & 15.80 & 79.40 & 45.40 \\
    CBP \cite{wang2019temporally} & 36.80 & 18.87 & 70.94 & 50.19 \\
    GDP \cite{chenrethinking} & 39.47 & 18.49 & - & - \\
    SCDM \cite{yuan2019semantic} & 54.44 & 33.43 & 74.43 & 58.08  \\
    \hline
    \textbf{CSMGAN} & \textbf{60.04} & \textbf{37.34} & \textbf{89.01} & \textbf{61.85}  \\ \hline
    \end{tabular}}
\end{table}

\begin{table}
    \small
    \centering
    \caption{Performance compared with previous methods on the DiDeMo dataset.}
    \label{tab:didemo}
    \setlength{\tabcolsep}{1.5mm}{
    \begin{tabular}{c|cccc}
    \hline \hline
    \multirow{2}*{Method} & R@1 & R@1 & R@5 & R@5 \\ 
    ~ & IoU=0.5 & IoU=0.7 & IoU=0.5 & IoU=0.7 \\ \hline
    MCN \cite{anne2017localizing} & 23.33 & 15.37 & 41.03 & 20.37 \\ 
    VSA-STV \cite{gao2017tall} & 25.38 & 14.49 & 68.56 & 26.92  \\
    VSA-RNN \cite{gao2017tall} & 24.94 & 14.52 & 68.39 & 26.10 \\
    CTRL  \cite{gao2017tall} & 26.45 & 15.36 & 68.78 & 28.43  \\
    ACRN \cite{liu2018attentive} & 27.44 & 16.65 & 69.43 & 29.45\\
    \hline
    \textbf{CSMGAN} & \textbf{29.44} & \textbf{19.16} & \textbf{70.77} & \textbf{41.61}  \\ \hline
    \end{tabular}}
\end{table}

\section{More Qualitative Results}
\label{sec:qualitative}
\begin{figure*}
    \centering
    \includegraphics[width=0.85\textwidth]{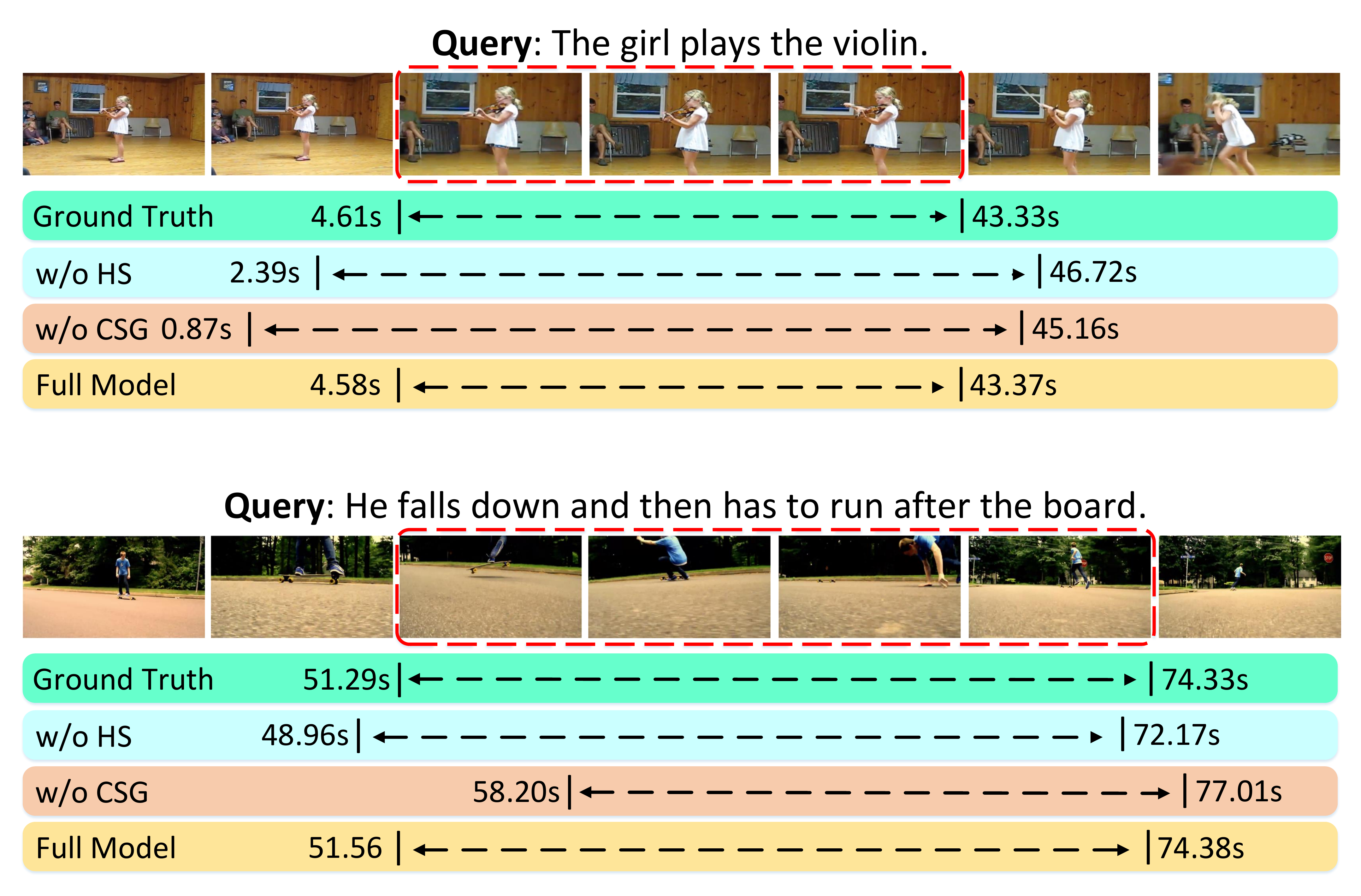}
    \caption{Qualitative visualization on Activity Caption dataset.}
    \label{fig:supp_acitivy}
\end{figure*}

\begin{figure*}
    \centering
    \includegraphics[width=0.85\textwidth]{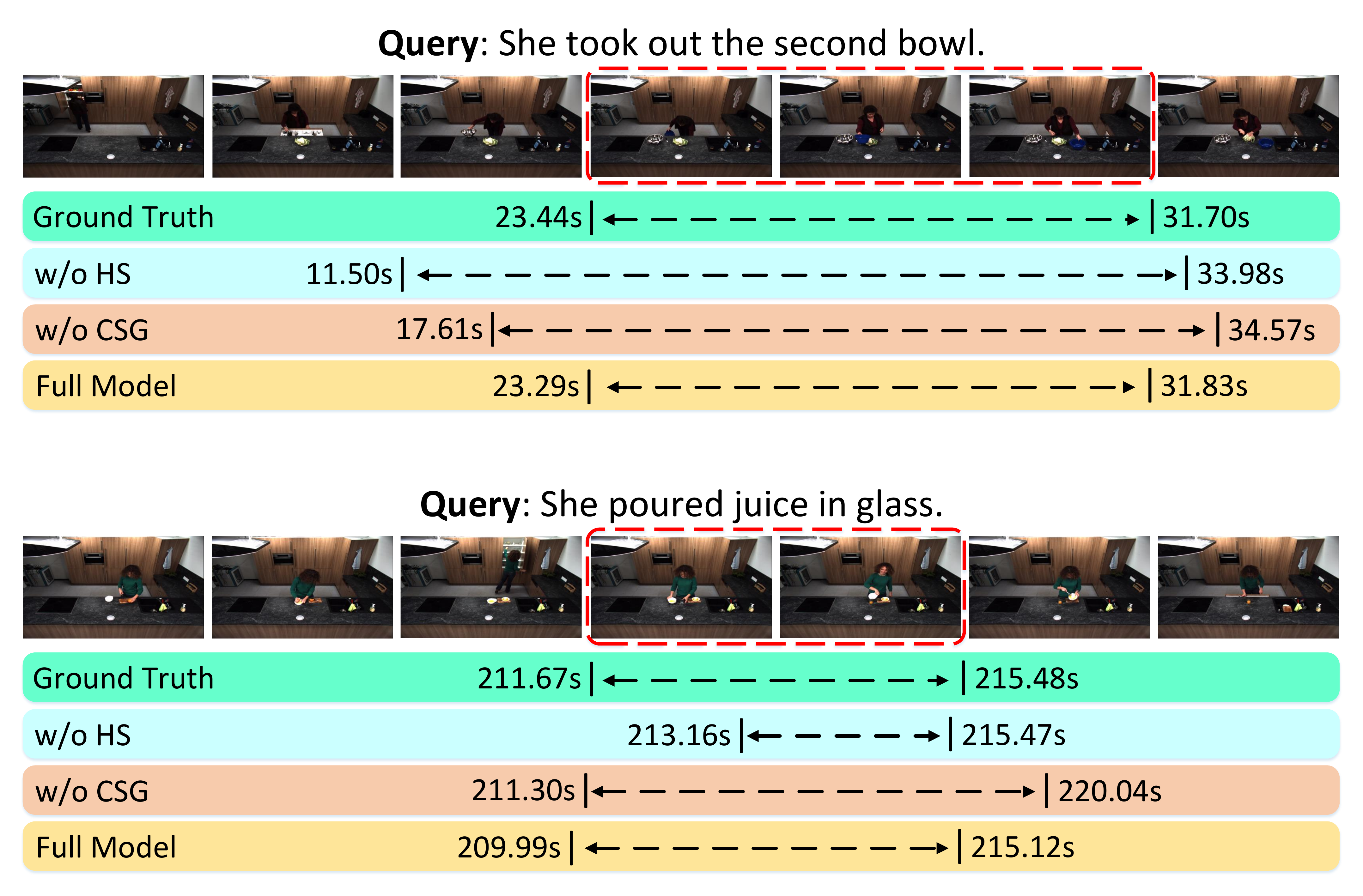}
    \caption{Qualitative visualization on TACoS dataset.}
    \label{fig:supp_tacos}
\end{figure*}

\begin{figure*}
    \centering
    \includegraphics[width=0.85\textwidth]{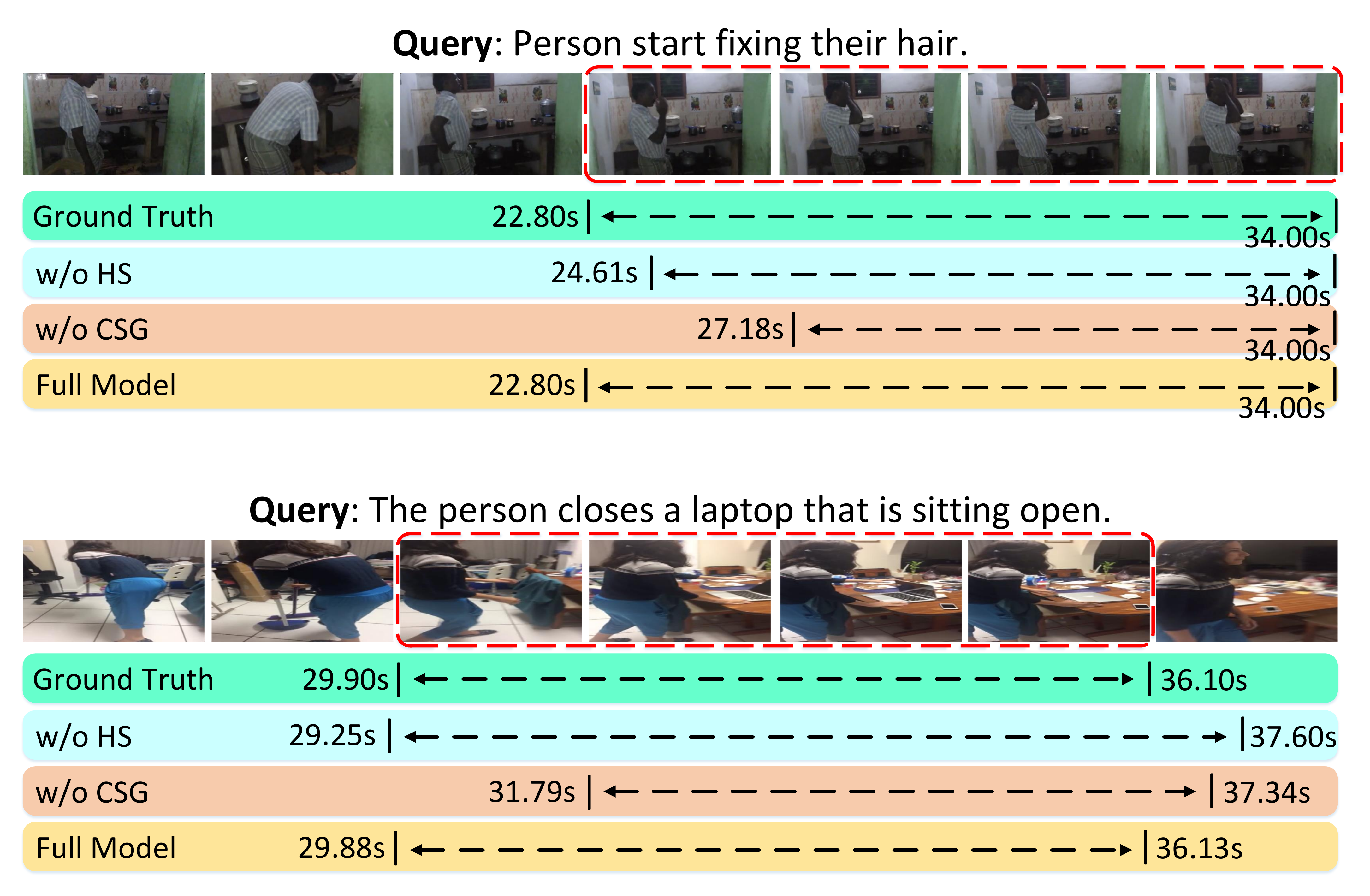}
    \caption{Qualitative visualization on Charades-STA dataset.}
    \label{fig:supp_charades}
\end{figure*}

\begin{figure*}
    \centering
    \includegraphics[width=0.85\textwidth]{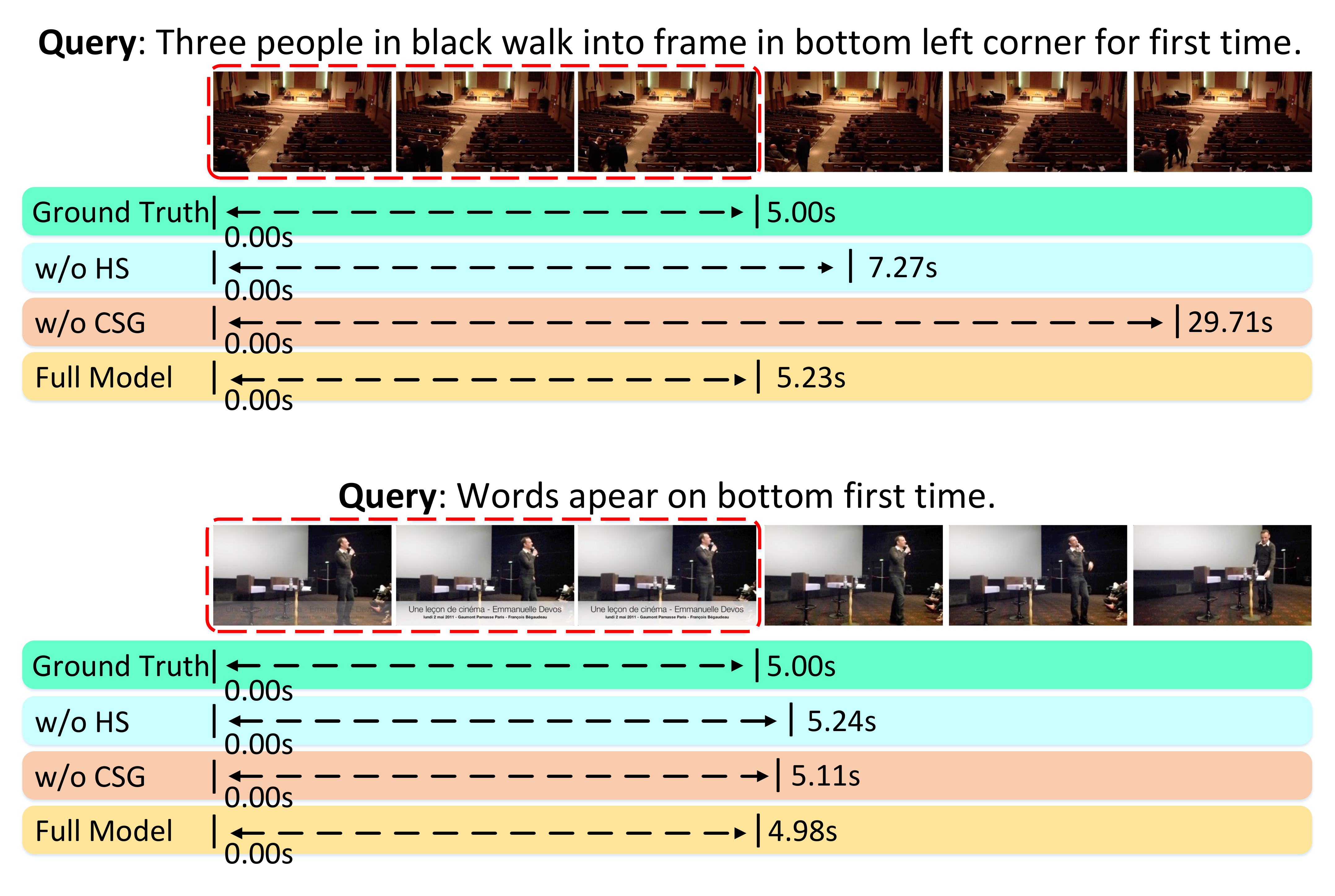}
    \caption{Qualitative visualization on DiDeMo dataset.}
    \label{fig:supp_didemo}
\end{figure*}

In this section, we provide more qualitative visualization results of our method on widely used datasets: 
\begin{itemize}
    \item Figure \ref{fig:supp_acitivy} shows results on Activity Caption \cite{caba2015activitynet} dataset.
    \item Figure \ref{fig:supp_tacos} shows results on TACoS \cite{regneri2013grounding} dataset.
    \item Figure \ref{fig:supp_charades} shows results on Charades-STA \cite{gao2017tall} dataset.
    \item Figure \ref{fig:supp_didemo} shows results on DiDeMo \cite{anne2017localizing} dataset.
\end{itemize}

% All videos contain the target objects among the frames. Therefore, how to accurately match the corresponding actions mentioned by the sentences is the key point in this task. 
For each dataset, we present four kinds of localization results on two sample videos, including three variants of our method and ground truth. For precise localization, our CSG (Cross- and Self-model Graph) mines the deep interaction between two modalities, and relates the instances within each modality. Meanwhile, HS (hierarchical structure) also contributes sentence understanding for better grounding. It is obvious that our full model achieves most precise localization result with the help of these two modules, which demonstrates the effectiveness of our proposed CSMGAN.

\end{document}